%% file: main.tex
\documentclass[pdflatex, sn-basic]{sn-jnl}

\usepackage{amsfonts}
\usepackage{comment}

\jyear{2021}%

\theoremstyle{thmstyleone}%
\newtheorem{theorem}{Theorem}
\newtheorem{definition}{Definition}

\newtheorem{corollary}{Corollary}
\newtheorem{proposition}{Proposition}

\input{commands}

\begin{document}

\title[PU-classification under class-prior shift]{Positive-Unlabeled Classification under Class-Prior Shift: A Prior-invariant Approach Based on Density Ratio Estimation}


\author*[1]{\fnm{Shota} \sur{Nakajima}}\email{nakajima@ms.k.u-tokyo.ac.jp}

\author[2,1]{\fnm{Masashi} \sur{Sugiyama}}\email{sugi@k.u-tokyo.ac.jp}

\affil*[1]{\orgname{The University of Tokyo}, \orgaddress{\state{Tokyo}, \country{Japan}}}

\affil[2]{\orgname{RIKEN}, \orgaddress{\state{Tokyo}, \country{Japan}}}



\abstract{\input{abstract}}




\maketitle

\section{Introduction}
\label{sec: introduction}
\input{introduction}

\section{Preliminaries}
\label{sec: relatedwork}
\input{relatedwork}

\section{Density ratio estimation for PU learning}
\label{sec: method}
\input{method}

\section{Discussions}
\label{sec: discussions}
\input{discussions}

\section{Experiments}
\label{sec: experiments}
\input{experiments}

\section{Conclusions}
\label{sec: conclusions}
\input{conclusions}

\section*{Acknowledgements}
MS was supported by KAKENHI 20H04206.

\begin{appendices}

\section{Proofs}
\label{sec: proofs}
\input{proofs}




\end{appendices}


\section*{Declarations}
\input{declarations}

\bibliography{references}


\end{document}

%% file: commands.tex
\newcommand{\abs}[1]{\left\lvert #1 \right\rvert}

\newcommand{\rmP}{\mathrm{P}}
\newcommand{\rmN}{\mathrm{N}}
\newcommand{\rmU}{\mathrm{U}}
\newcommand{\expect}[1]{\mathbb{E} \left[ #1 \right]}
\newcommand{\expectm}[2]{\mathbb{E}_{#1} \left[ #2 \right]}
\newcommand{\expectP}[1]{\expectm{\rmP}{#1}}
\newcommand{\expectN}[1]{\expectm{\rmN}{#1}}
\newcommand{\expectU}[1]{\expectm{\rmU}{#1}}
\newcommand{\empexpectm}[2]{\widehat{\mathbb{E}}_{#1} \left[ #2 \right]}
\newcommand{\empexpectP}[1]{\empexpectm{\rmP}{#1}}

\newcommand{\empexpectU}[1]{\empexpectm{\rmU}{#1}}
\newcommand{\numP}{n_\rmP}

\newcommand{\numU}{n_\rmU}

\newcommand{\numUs}{\numU^\prime}
\newcommand{\setP}{\mathcal{X}_\rmP}

\newcommand{\setU}{\mathcal{X}_\rmU}

\newcommand{\setUs}{\mathcal{X}_\rmU^\prime}
\newcommand{\sign}{\mathrm{sign}}

\newcommand{\BR}[3]{\mathrm{BR}_{#1} \left( {#2} \parallel {#3} \right)}
\newcommand{\pri}{\pi}
\newcommand{\pris}{\pri^\prime}
\newcommand{\cost}{c}
\newcommand{\costs}{\cost^\prime}

\newcommand{\di}[1]{#1^\prime}
\newcommand{\ddi}[1]{#1^{\prime \prime}}
\newcommand{\Radem}[2]{\mathfrak{R}_{#1}^{#2}}
\newcommand{\AUC}{\mathrm{AUC}}
\newcommand{\indi}[1]{1\{#1\}}
\newcommand{\avg}[2]{$#1 \pm #2$}
\newcommand{\avgbold}[2]{$\mathbf{#1 \pm #2}$}
\newcommand{\avgtwo}[2]{
    \begin{tabular}{r}
        $#1$ \\
        $\pm #2$
    \end{tabular}
}
\newcommand{\avgboldtwo}[2]{
    \begin{tabular}{r}
        $\mathbf{#1}$ \\
        $\pm \mathbf{#2}$
    \end{tabular}
}

%% file: abstract.tex

Learning from positive and unlabeled (PU) data is an important problem in various applications. Most of the recent approaches for PU classification assume that the class-prior (the ratio of positive samples) in the training unlabeled dataset is identical to that of the test data, which does not hold in many practical cases. In addition, we usually do not know the class-priors of the training and test data, thus we have no clue on how to train a classifier without them. To address these problems, we propose a novel PU classification method based on density ratio estimation. A notable advantage of our proposed method is that it does not require the class-priors in the training phase; class-prior shift is incorporated only in the test phase. We theoretically justify our proposed method and experimentally demonstrate its effectiveness.

%% file: introduction.tex

Positive-unlabeled (PU) classification is a problem of training a binary classifier from only positive and unlabeled data \citep{Letouzey2000,Elkan2008}. This problem is important when it is difficult to gather negative data, and appears in many applications, such as inlier-based outlier detection \citep{Blanchard2010}, land-cover classification \citep{Li2011}, matrix completion \citep{Hsieh2015}, and sequential data classification \citep{Li2009,Nguyen2011}. Several heuristic approaches for PU classification have been proposed in the past \citep{Liu2003,Li2003}, which aim to identify negative samples in the unlabeled dataset, yet they heavily rely on the heuristic strategy and data separability assumption. One of the most theoretically and practically effective methods for PU classification was established by \citet{DuPlessis2014pu,DuPlessis2015}, called \emph{unbiased PU classification}. It rewrites the classification risk in terms of the distributions over positive and unlabeled samples, and obtains an unbiased estimator of the risk without negative samples. Although unbiased PU classification works well with simple models such as linear-in-parameter models, it easily suffers from overfitting with flexible models such as deep neural networks. To overcome this problem, a \emph{non-negative risk estimator} \citep{Kiryo2017} for PU classification was proposed.
\par
Besides unbiased PU classification, various approaches for PU classification have also been proposed recently. For example, generative adversarial networks (GAN) have been applied to PU classification by \citet{Hou2018}, which allows one to learn from a small number of positive samples. \citet{Zhang2019} introduced \emph{ambiguity} to unlabeled samples and performed PU label disambiguation (PULD) based on margin maximization to determine the true labels of all unlabeled examples. A variational approach and a data augmentation method based on \emph{Mixup} \citep{Zhang2018} were proposed by \citet{Chen2020} for PU classification without explicit estimation of the class-prior of the training data.
\par
One of the drawbacks of these approaches is that the distribution of the test data must be identical to that of the training data, which may be violated in practice \citep{Quionero2009}. For example, the class-prior (the ratio of positive data) in the training unlabeled dataset might be different from that of the test data, known as the \emph{class-prior shift} problem. To cope with this problem, \citet{Charoenphakdee2019} showed that classification under class-prior shift can be written as cost-sensitive classification, and proposed a risk-minimization approach and a density ratio estimation \citep{Sugiyama2012} approach. In their study, both the class-priors of the training and test data are assumed to be given in advance, but this assumption does not hold in many practical cases. Therefore, we need to estimate them with the training and test data. 
\par 
However, it is usually hard to obtain samples from the test distribution at the training time, and this is not natural because we do not know whether the prior shift would occur or not in advance. Furthermore, the training data would be inaccessible once the training has been completed, especially in privacy-concerned situations such as click analysis \citep{McMahan2013}, purchase prediction \citep{Martinez2018}, and voting prediction \citep{Coletto2015}. In that kind of problem, the model is trained with data including personal information, and only the trained model is kept while the data must be discarded. This implies that we are not allowed to use training data when a classifier is adapted to an unknown test distribution.

\begin{table}[ht]
    \centering
    \caption{Comparisons of representative existing PU classification methods. uPU was proposed by \citet{DuPlessis2014pu,DuPlessis2015}, nnPU was proposed by \citet{Kiryo2017}, GenPU was proposed by \citet{Hou2018}, PULD was proposed by \citet{Zhang2019}, VPU was proposed by \citet{Chen2020}, and PUa was proposed by \citet{Charoenphakdee2019}.}
    \label{tab: existing_works}
    \setlength{\tabcolsep}{5.0pt}
    \begin{tabular*}{\linewidth}{@{\extracolsep{\fill}}lccccccc}
        \toprule
        & uPU & nnPU & GenPU & PULD & VPU & PUa & Ours \\
        \midrule
        \begin{tabular}{l}
            Excess risk bound and its \\
            convergence rate analysis
        \end{tabular} & $\checkmark$ & $\checkmark$ & $\times$ & $\checkmark$ & $\times$ & $\times$ & $\checkmark$ \\ \midrule
        \begin{tabular}{l}
            Learning a classifier without \\
            given class-prior(s)
        \end{tabular} & $\times$ & $\times$ & $\times$ & $\times$ & $\checkmark$ & $\times$ & $\checkmark$ \\ \midrule
        \begin{tabular}{l}
            Adaptable to test-time \\
            class-prior shift
        \end{tabular} & $\times$ & $\times$ & $\times$ & $\times$ & $\times$ & $\checkmark$ & $\checkmark$ \\
        \bottomrule
    \end{tabular*}
\end{table}

\par
To overcome these problems, we propose an approach based on density ratio estimation \citep{Sugiyama2012}. Density ratio estimation for PU classification has appeared in several existing works \citep{Charoenphakdee2019,Kato2019}, yet their studies have no guarantees on the theoretical relationship between binary classification and density ratio estimation. Our proposed method can train a classifier without given knowledge of the class-priors, and adapt to the test-time class-prior shift without the training data. Table \ref{tab: existing_works} summarizes comparisons of representative existing methods and our proposed method. Our main contributions are: (i) We propose a method for PU classification under test-time class-prior shift with unknown class-priors, (ii) We theoretically justify the proposed method, and (iii) Experimental results show the effectiveness of the proposed method.

%% file: relatedwork.tex
In this section, we introduce the notations, and review the concepts of unbiased/non-negative PU classification, cost-sensitive classification, and density ratio estimation.

\subsection{Problem formulation} 
Let $X \in \mathbb{R}^d$ and $Y \in \{\pm 1\}$ be the input and output random variables, where $d$ denotes the dimensionality of the input variable. Let $p(x, y)$ be the \emph{underlying joint density} of $(X, Y)$ and $p(x)$ be the input marginal density. We denote the positive and negative class-conditional densities as
\begin{equation}
    \begin{split}
        p_+(x)&=p(x \mid Y=+1) \\
        p_-(x)&=p(x \mid Y=-1).
    \end{split}
\end{equation}
Let $\pri = p(Y=+1)$ be the positive \emph{class-prior probability}. Assume we have i.i.d.~sample sets $\setP$ and $\setU$ from $p_+(x)$ and $p(x)$ respectively, and let $\numP = \abs{\setP}$ and $\numU = \abs{\setU}$, where $\abs{\cdot}$ denotes the cardinality of a set. We denote the expectations over each class-conditional density as
\begin{equation}
    \begin{split}
        &\expectP{\cdot} = \expectm{X \sim p_+}{\cdot} \\
        &\expectN{\cdot} = \expectm{X \sim p_-}{\cdot} \\
        &\expectU{\cdot} = \expectm{X \sim p}{\cdot} = \expectm{X}{\cdot},
    \end{split}
\end{equation}
and their empirical counterparts as 
\begin{equation}
    \begin{split}
        &\empexpectP{f(X)} = \frac{1}{\numP} \sum_{x \in \setP} f(x) \\
        &\empexpectU{f(X)} = \frac{1}{\numU} \sum_{x \in \setU} f(x),
    \end{split}
\end{equation}
where $f$ is an arbitrary function of $x \in \mathbb{R}^d$. Let $g : \mathbb{R}^d \to \mathbb{R}$ be a real-valued decision function. The purpose of binary classification is to minimize the expected classification risk
\begin{equation}
    R(g) = \expectm{X, Y}{\indi{\sign(g(X)) \neq Y}},
\end{equation}
where $1\{\cdot\}$ denotes the indicator function. Since the optimization problem based on the zero-one loss is computationally infeasible \citep{Arora1997,Bartlett2006}, a surrogate loss function $\ell : \{\pm 1\} \times \mathbb{R} \to \mathbb{R}$ is used in practice. Classification risk with respect to surrogate loss is defined as
\begin{equation}
    R_\ell(g) = \expectm{X, Y}{\ell(Y, g(X))}.
\end{equation}


\subsection{Unbiased/non-negative PU classification}
The surrogate classification risk can be written as
\begin{equation}
    R_\ell(g) = \pri \expectP{\ell(+1, g(X))} + (1 - \pri) \expectN{\ell(-1, g(X))}.
\end{equation}
Since negative samples are unavailable in PU classification, we rewrite the expectation over the negative class-conditional distribution as 
\begin{equation}
    (1 - \pri) \expectN{\ell(-1, g(X))} = \expectU{\ell(-1, g(X))} - \pri \expectP{\ell(-1, g(X))},
\end{equation}
where $p(x) = \pri p_+(x) + (1 - \pri) p_-(x)$ is used \citep{DuPlessis2014pu}. Then, the risk can be approximated directly with $\setP$ and $\setU$ as
\begin{equation} \label{eq: PUrisk}
    \widehat{R}_\ell(g) = \pri \empexpectP{\ell(+1, g(X))} - \pri \empexpectP{\ell(-1, g(X))} + \empexpectU{\ell(-1, g(X))}.
\end{equation}
The \emph{empirical risk estimator} $\widehat{R}_\ell(g)$ is unbiased and consistent \citep{Niu2015}, i.e., $\expect{\widehat{R}_\ell(g)} = R_\ell(g)$ where the expectation $\mathbb{E}$ is taken over all of the samples, and $\widehat{R}_\ell(g) \to R_\ell(g)$ as $\numP, \numU \to \infty$. 
\par
Unbiased PU classification easily suffers from overfitting when we use a flexible model such as neural networks, because the model can be so powerful that it fits all of the given samples, and then the empirical risk goes negative \citep{Kiryo2017}. To mitigate this problem, a non-negative risk correction approach was proposed \citep{Kiryo2017}. Since 
\begin{equation}
    \expectU{\ell(-1, g(X))} - \pri \expectP{\ell(-1, g(X))} = (1 - \pri)\expectN{\ell(-1, g(X))} \geq 0
\end{equation}
holds for any non-negative loss function, we correct the corresponding part of the expected risk to be non-negative. Approximating the expectations by sample averages gives the non-negative risk estimator:
\begin{equation}
    \widetilde{R}_\ell(g) = \pri \empexpectP{\ell(+1, g(X))} + \left( \empexpectU{\ell(-1, g(X))} - \pri \empexpectP{\ell(-1, g(X))} \right)_+,
\end{equation}
where $(\cdot)_+ = \max(0, \cdot)$. The non-negative risk estimator is biased yet consistent, and its bias decreases exponentially with respect to $\numP + \numU$ \citep{Kiryo2017}.


\subsection{Cost-sensitive classification} \label{subsec: costsensitive}
For arbitrary \emph{false-positive cost} parameter $c \in (0, 1)$, cost-sensitive classification is defined as a problem of minimize the following risk \citep{Elkan2001,Scott2012} :
\begin{equation}
    R_{\pri, \cost}(g) = (1 - \cost) \pri \expectP{\indi{\mathrm{sign}(g(X)) \neq +1}} + \cost (1 - \pri) \expectN{ \indi{\mathrm{sign}(g(X)) \neq -1}}.
\end{equation}
When $\cost = 1/2$, cost-sensitive classification reduces to ordinary binary classification, up to unessential scaling factor $1/2$. \citep{Charoenphakdee2019} showed that classification under class-prior shift can be formulated as cost-sensitive classification. For example, let $\pris \in (0, 1)$ be the class-prior of the test distribution, then $R_{\pris, 1/2} \propto R_{\pri, \cost}$ with $\cost = \frac{\pri(1 - \pris)}{\pri(1 - \pris) + (1 - \pri)\pris}$. 


\subsection{Class-prior estimation}
In unbiased/non-negative PU classification, the class-prior is assumed to be given, which does not hold in many practical cases. Unfortunately, we cannot treat $\pri$ as a hyperparameter to be tuned, because there exists a trivial solution such as $\pri = 0$ and $g(x) \equiv \mathrm{argmin}_v \ell(-1, v)$. One of the solutions to this problem is to estimate both the training and test class-priors by existing methods respectively with positive, training-unlabeled, and test-unlabeled datasets. In fact, it is known that class-prior estimation is an ill-posed problem, without any additional assumptions \citep{Blanchard2010,Blanchard2013}. For example, if 
\begin{equation}
    p(x) = \kappa p_+(x) + (1 - \kappa) p_-(x)
\end{equation}
holds, then there exists a density $p_-^\prime(x)$ such that 
\begin{equation}
    p(x) = (\kappa - \delta)p_+(x) + (1 - \kappa + \delta)p_-^\prime(x)
\end{equation}
for $0 \leq \delta \leq \kappa$. In practice, an alternative goal of estimating the maximum mixture proportion 
\begin{equation}
    \small
    \kappa^* = \max \{ \kappa \in [0, 1] : \exists p_- \ \mathrm{s.t.} \ p(x) = \kappa p_+(x) + (1 - \kappa)p_-(x) \}
\end{equation}
is pursued \citep{Blanchard2010,Blanchard2013}. The \emph{irreducibility assumption} \citep{Blanchard2010,Blanchard2013} gives a constraint on the true underlying densities which ensures that $\kappa^*$ is the unique solution of prior estimation.
\begin{definition}[Irreducibility \citep{Blanchard2010,Blanchard2013}] \label{def: irreducibility}
    Let $G$ and $H$ be probability distributions on $(\mathbb{R}^d, \mathfrak{S})$ where $\mathfrak{S}$ is a Borel algebra on $\mathbb{R}^d$. We say that $G$ is irreducible with respect to $H$, if there is no decomposition of the form $G = \kappa H + (1 - \kappa) H^\prime $ where $H^\prime$ is some probability distribution and $0 < \kappa \leq 1$.
\end{definition}
Let $P$, $P_+$, and $P_-$ be the cumulative distribution functions of $p$, $p_+$, and $p_-$ respectively. Under the irreducibility assumption, the class-prior is identical to the maximum mixture proportion.
\begin{proposition} [\citep{Blanchard2010,Blanchard2013}] \label{thm: maximum_mixture}
    Let $P$, $P_+$, and $P_-$ be probability distributions on $(\mathbb{R}^d, \mathfrak{S})$. If $P = \pri P_+ + (1 - \pri) P_-$ and $P_-$ is irreducible with respect to $P_+$, then 
    \begin{equation}
        \begin{split}
            \pri &= \max\{\kappa \in [0, 1] : \exists Q \ \mathrm{s.t.} \ P = \kappa P_+ + (1 - \kappa)Q \} \\
            &= \inf_{S \in \mathfrak{S}, P_+(S) > 0} \frac{P(S)}{P_+(S)}.
        \end{split}
    \end{equation}
\end{proposition}
Note that the set $S \in \mathfrak{S}$ corresponds to a measurable hypothesis $h : \mathbb{R}^d \to \{-1, +1\}$ bijectively.
Based on these facts, several works for class-prior estimation have been proposed \citep{Blanchard2010,Blanchard2013,Scott2015,Ramaswamy2016,DuPlessis2017}.
However, they usually work with kernel methods which are computationally hard to apply to large-scale and high-dimensional data. Furthermore, since the unbiased/non-negative risk estimators depend on the class-prior, an estimation error of the class-prior directly affects the optimization. In addition, we usually do not have a sample set from the test distribution at the training time, and thus cannot even estimate the class-prior of the test data by such existing methods.


\subsection{Density ratio estimation}
The ratio of two probability densities has attracted attention in various problems \citep{Sugiyama2009,Sugiyama2012}. Density ratio estimation (DRE) \citep{Sugiyama2012} aims to directly estimate the ratio of two probabilities, instead of estimating the two densities separately. \citet{Sugiyama2011} showed that various existing DRE methods \citep{Sugiyama2008, Kanamori2009, Kato2019} can be unified from the viewpoint of Bregman divergence minimization, so we consider the DRE problem as a Bregman divergence minimization problem.
\par
Here we consider estimating the ratio of the positive class-conditional density to the input marginal density. Let $r^*(x) = p_+(x)/p(x)$ be the true density ratio and $r : \mathbb{R}^d \to [0, \infty)$ be a density ratio model. For a convex and differentiable function $f : [0, \infty) \to \mathbb{R}$, the expected Bregman divergence, which measures the discrepancy from $r^*$ to $r$, is defined as
\begin{equation} \label{eq: bregman}
    \begin{split}
        \BR{f}{r^*}{r} &= \int \left( f(r^*(x)) - f(r(x)) - \di{f}(r(x))(r^*(x)-r(x)) \right)p(x)\mathrm{d}x \\
        &= \expectP{-\di{f}(r(X))} + \expectU{\di{f}(r(X))r(X) - f(r(X))} + \mathrm{const.},
    \end{split}
\end{equation}
where the constant term does not include $r$. The function $f$ is called the generator function of the Bregman divergence \citep{Menon2016dr}. We can see that the Bregman divergence of DRE does not contain the class-prior $\pri$, and can be approximated by taking empirical averages over the positive and unlabeled datasets, except the constant term.
\par 
Similarly to the case of unbiased PU classification, it was revealed that empirical Bregman divergence minimization often suffers from severe overfitting when we use a highly flexible model \citep{Kato2020}. To mitigate this problem, non-negative risk correction for the Bregman divergence minimization was proposed, based on the idea of non-negative PU classification \citep{Kiryo2017}. 
\par
The objective function for Bregman divergence minimization is defined by Eq. (\ref{eq: bregman}) without the constant term
\begin{equation}
    \mathcal{L}_f(r) = \expectP{-\di{f}(r(X))} + \expectU{\di{f}(r(X))r(X) - f(r(X))}.
\end{equation}
We also consider its empirical counterpart $\widehat{\mathcal{L}}_f(r)$. Let us denote
\begin{equation}
    \begin{split}
        f^*(t) &= t\di{f}(t) - f(t) \\
        \mathfrak{F}(t) &= f^*(t) - f^*(0),
    \end{split}
\end{equation}
then $\mathfrak{F}$ is non-negative on $[0, \infty)$ because $\di{(f^*)}(t) = \ddi{f}(t) \geq 0$ (i.e., $f$ is convex.). We pick a lower bound of $\pri$ as $\alpha$ and then we have
\begin{equation}
        \expectU{\mathfrak{F}(r(X))} - \alpha \expectP{\mathfrak{F}(r(X))} \geq  (1 - \pri) \expectN{\mathfrak{F}(r(X))} \geq 0.
\end{equation}
Thus, we define the corrected empirical estimator of $\mathcal{L}$ as
\begin{equation} \label{eq: nnBR}
    \begin{split}
        \widetilde{\mathcal{L}}_f(r) &= \empexpectP{-\di{f}(r(X)) + \alpha \mathfrak{F}(r(X))} \\
        & \quad + \left( \empexpectU{\mathfrak{F}(r(X))} - \alpha \empexpectP{\mathfrak{F}(r(X))} \right)_+ + f^*(0),
    \end{split}
\end{equation}
where $(\cdot)_+ = \max(0, \cdot)$. $\widetilde{\mathcal{L}}_f$ is consistent as long as $0 \leq \alpha \leq \pi$, and its bias decreases exponentially with respect to $\numP + \numU$. Even though we do not have any knowledge of $\pri$, we can tune $\alpha$ as a hyperparameter to minimize the empirical estimator without non-negative correction $\widehat{\mathcal{L}}_f(r)$, which contains neither $\pri$ nor $\alpha$, with the positive and unlabeled validation datasets.

%% file: method.tex

In this section, we formulate a cost-sensitive binary classification problem as a density ratio estimation problem, and propose a method of Density Ratio estimation for PU learning (DRPU). All proofs are given in Appendix \ref{sec: proofs}.

\subsection{Excess risk bounds} \label{subsec: dre}
From Bayes' rule, we have $p(Y=+1 \mid X) = \pri p_+(x)/p(x) = \pri r^*(x)$. Therefore, the optimal solution of the Bregman divergence minimization, $r = r^*$ gives a Bayes optimal classifier by thresholding $p(Y=+1 \mid X) = 1/2$, and it motivates us to use $r$ for binary classification. However, this statement only covers the optimal solution, and it is unclear how the classification risk grows along with the Bregman divergence.
To cope with this problem, we interpret the DRE as minimization of an upper bound of the excess classification risk. Although the relationship between DRE and class probability estimation has been studied by \citet{Menon2016dr}, differently from that, our work focuses on the ratio of the densities of positive and unlabeled data, and the value of the Bregman divergence does not depend on the class-prior $\pri$. 
\par 
We denote the \emph{Bayes optimal risk} as $R_{\pri, \cost}^* = \inf_{g \in \mathcal{F}} R_{\pri, \cost}(g)$, where $\mathcal{F}$ is the set of all measurable functions from $\mathbb{R}^d$ to $\mathbb{R}$, and the difference $R_{\pri, \cost(g)} - R_{\pri, \cost}^*$ is called the \emph{excess risk} for $R_{\pri, \cost}$. The following theorem associates DRE with cost-sensitive classification under a strong convexity assumption on $f$, justifying solving binary classification by DRE.
\begin{theorem} \label{thm: classification_ratio}
    Let $f$ be a $\mu$-strongly convex function, i.e., $\mu = \inf_{t \in [0, \infty)}\ddi{f}(t) > 0$. Then, for any $\pri \in (0, 1)$, $\cost \in (0, 1)$, $r : \mathbb{R}^d \to [0, \infty)$, and $h_\cost = \sign(\pri r - \cost)$, we have
    \begin{equation}
        R_{\pri, \cost}(h_\cost) - R_{\pri, \cost}^* \leq \pri \sqrt{\frac{2}{\mu} \BR{f}{r^*}{r}}.
    \end{equation}
\end{theorem}
As we have already seen in Section \ref{subsec: costsensitive}, the class-prior shift problem can be transformed into a cost-sensitive classification problem. Next, we extend the excess risk bound in Theorem \ref{thm: classification_ratio} to the case of prior shift. Let $\pris$ be the test-time class-prior and $\costs$ be the test-time false positive cost. Note that $\costs = 1/2$ corresponds to standard binary classification. The classification risk with respect to the test distribution is defined as
\begin{equation}
    \begin{split}
        R_{\pris, \costs}(g) &= (1 - \costs)\pris \expectP{\indi{\sign(g(X)) \neq +1 }} \\
        & \quad + \costs(1 - \pris) \expectN{\indi{\sign(g(X)) \neq -1 }}.
    \end{split}
\end{equation}
The following theorem gives an excess risk bound for $R_{\pris, \costs}$.
\begin{theorem} \label{thm: classification_shift}
    Let $f$ be a $\mu$-strongly convex function. Then, for any $\pri, \pris \in (0, 1)$, $\costs \in (0, 1)$, $r : \mathbb{R}^d \to [0, \infty)$, and $h_{\cost_0} = \sign(\pri r - c_0)$, we have
    \begin{equation}
        R_{\pris, \costs}(h_{\cost_0}) - R_{\pris, \costs}^* \leq C \sqrt{\frac{2}{\mu}\BR{f}{r^*}{r}},
    \end{equation}
    where $\cost_0 = \frac{\costs \pri (1 - \pris)}{(1 - \costs)(1 - \pri)\pris + \costs \pri (1 - \pris)}$ and $C = \pri \frac{\costs + \pris - 2\costs \pris}{\cost_0 + \pri - 2\cost_0 \pri}$.
\end{theorem}
Note that this is a generalized version of Theorem \ref{thm: classification_ratio}, which is the case of $\pris = \pri$ and $\costs = \cost$. This result shows that even when the class-prior and cost are shifted, by changing the classification threshold to $\cost_0$, the classification risk can still be bounded by the Bregman divergence of DRE.


\subsection{Estimating the class-priors} \label{subsec: prior}
Although we can train a model $r$ and deal with a prior shift problem without the knowledge of the class-priors, we still need to estimate them to determine the classification threshold $\cost_0 / \pri$. Based on Proposition \ref{thm: maximum_mixture}, we propose the following estimator of $\pri$:
\begin{equation} \label{eq: priorestimator}
    \hat{\pri}(r) = \inf_{h \in \mathcal{H}_r} \frac{\widehat{P}(h)}{\widehat{P}_+(h)},
\end{equation}
where 
\begin{equation}
    \begin{split}
        \widehat{P}(h) &= \empexpectU{1\{h(X) = +1\}} \\
        \widehat{P}_+(h) &= \empexpectP{1\{h(X) = +1\}}
    \end{split}
\end{equation}
and
\begin{equation}
    \mathcal{H}_r = \big\{ h : \mathbb{R}^d \to \{\pm 1\} \mid \exists \theta \in \mathbb{R}, h(x) = \sign(r(x) - \theta) \land \widehat{P}_+(h) > \bar{\gamma} \big\}.
\end{equation}
Here, 
\begin{equation}
    \bar{\gamma} = \frac{1}{\gamma} \mathrm{max} (\varepsilon(\numP, 1/\numP), \varepsilon(\numU, 1/\numU))
\end{equation}
and,
\begin{equation}
    \varepsilon(n, \delta) = \sqrt{\frac{4 \log (en / 2)}{n}} + \sqrt{\frac{\log (2 / \delta)}{2n}}
\end{equation}
for some $n > 0$ and $0 < \delta < 1$, and $0 < \gamma < 1$ is an arbitrarily fixed constant. The main difference from the estimator proposed by \citet{Blanchard2010} and \citet{Blanchard2013} is that the hypothesis $h$ is determined by thresholding the trained density ratio model $r$, thus we need no additional training of the model.
\par 
To consider convergence of the proposed estimator, we introduce the concept of the \emph{Area Under the receiver operating characteristic Curve (AUC)}, which is a criterion to measure the performance of a score function for bipartite ranking \citep{Menon2016br}. For any real-valued score function $s : \mathbb{R}^d \to \mathbb{R}$, AUC is defined as
\begin{equation} \label{eq: auc}
    \mathrm{AUC} (s) = \expect{\indi{(Y - Y^\prime)(s(X) - s(X^\prime)) > 0} \mid Y \neq Y^\prime},
\end{equation}
where the expectation is taken over $X$, $X^\prime$, $Y$, and $Y^\prime$. In addition, we define the AUC risk and the optimal AUC risk as $R_\AUC(r) = 1 - \AUC(r)$ and $R_\AUC^* = \inf_{s \in \mathcal{F}}R_\AUC(s)$, where $\mathcal{F}$ is a set of all measurable functions from $\mathbb{R}^d$ to $\mathbb{R}$. Then, the following theorem gives a convergence guarantee of the estimator. 
\begin{theorem} \label{thm: priorestimator}
    For $\hat{\pri}(r)$ defined by Eq. (\ref{eq: priorestimator}), with probability at least $(1 - 1/\numP)(1 - 1/\numU)$, we have
    \begin{equation}
        \abs{\hat{\pri}(r) - \pri} \leq \xi \left( R_\AUC(r) - R_\AUC^* \right) + \mathcal{O} \left( \sqrt{\frac{\log \numP}{\numP}} + \sqrt{\frac{\log \numU}{\numU}} \right).
    \end{equation}
    Here, $\xi$ is an increasing function such that 
    \begin{equation}
        \xi \left( R_\AUC(r) - R_\AUC^* \right) \leq \frac{2(1 - \pri)}{1 - \bar{\gamma}^2} R_\AUC(r),
    \end{equation}
    and $\xi(0) \to 0$ as $\bar{\gamma} \to 0$.
\end{theorem}
This result shows that a \emph{better score function} in the sense of AUC tends to result in a \emph{better estimation} of $\pri$. Furthermore, we can see that the scale of $r$ is not important for the estimator $\hat{\pri}(r)$; it just needs to be a good score function, therefore $r$ can be used not only to estimate $\pri$ but also to estimate $\pris$. Given a sample set $\setUs$ from the test density $p^\prime(x) = \pris p_+(x) + (1 - \pris) p_-(x)$, we propose the following estimator of the test prior $\pris$:
\begin{equation} \label{eq: priorestimator_shift}
    \hat{\pri}^\prime(r) = \inf_{h \in \mathcal{H}_r} \frac{\widehat{P}^\prime(h)}{\widehat{P}_+(h)},
\end{equation}
where $\widehat{P}^\prime(h) = \empexpectm{\rmU^\prime}{1\{h(X) = +1\}}$. Replacing $\pri$ by $\pris$, $\hat{\pri}$ by $\hat{\pri}^\prime$, and $\numU$ by $\numUs = \abs{\setUs}$ in Theorem \ref{thm: priorestimator} , we can obtain an error bound of $\hat{\pri}^\prime$.
\par
In Eq. (\ref{eq: priorestimator_shift}), we require the dataset from the positive class-conditional distribution and the test-time input marginal distribution. As described in Sections \ref{sec: introduction} and \ref{sec: relatedwork}, we sometimes do not have access to the training data at the test-time. Fortunately in Eq. (\ref{eq: priorestimator_shift}), we need only the value of $\widehat{P}_+(h)$ for each $h$, and we do not care about the samples themselves. Also, $\widehat{P}_+(h)$ takes ascending piece-wise constant values from 0 to 1 with interval $1/\numP$. Hence, preserving the list of intervals $\{\Theta_i\}_{i=0}^{\numP}$ such that $\widehat{P}_+(\sign(r(X)-\theta)) = i/\numP$ for all $\theta \in \Theta_i$ at the training time, we can use it to estimate the test-time class-prior, without preserving the training data themselves.


\subsection{Practical implementation} \label{subsec: implementation}
The entire flow of our proposed method is described in Algorithm \ref{alg: DRPU}. Since the strong convexity of the generator $f$ of the Bregman divergence is desired, we may employ the quadratic function $f(t) = t^2 / 2$. In this case, the DRE method is called \emph{Least-Square Importance Fitting (LSIF)} \citep{Kanamori2009}. As a parametric model $r$, a deep neural network may be used and optimized by stochastic gradient descent. Details of the stochastic optimization method for $\widetilde{\mathcal{L}}_f$ are described in Algorithm \ref{alg: nnBR}. In the prior estimation step, it is recommended to use data that is not used in the training step to avoid overfitting, especially when we are using flexible models. So we split the given data into the training and validation sets, then use the validation set to tune hyperparameters and estimate the class-priors. 
\begin{algorithm}
    \caption{DRPU}
    \label{alg: DRPU}
    \begin{algorithmic}[1]
        \Require Training datasets $(\setP, \setU)$, test dataset $\setUs$, 
        \Ensure Classifier $h : \mathbb{R}^d \to \{-1, +1\}$
        \State Split $(\setP, \setU)$ into training set $(\setP^{\mathrm{tr}} \setU^{\mathrm{tr}})$ and validation set $(\setP^{\mathrm{val}}, \setU^{\mathrm{val}})$
        \While {no stopping criterion has been met}
            \State Optimize $r$ with $(\setP^{\mathrm{tr}}, \setU^{\mathrm{tr}})$ by minimizing $\widetilde{\mathcal{L}}_f$
        \EndWhile
        \State Estimate $\hat{\pri}$ with $r$ and $(\setP^{\mathrm{val}}, \setU^{\mathrm{val}})$ by Eq. (\ref{eq: priorestimator})
        \State Preserve the list of intervals $\{\Theta_i\}_{i=0}^{\numP}$
        \State // \emph{Obtain $\setUs$ at the test-time}
        \State Estimate $\hat{\pri}^\prime$ with $r$, $\setUs$, and $\{\Theta_i\}_{i=0}^{\numP}$ by Eq. (\ref{eq: priorestimator_shift})
        \State Determine $\hat{\cost}_0 = \frac{\costs \hat{\pri} (1 - \hat{\pri}^\prime)}{(1 - \costs)(1 - \hat{\pri})\hat{\pri}^\prime + \costs \hat{\pri} (1 - \hat{\pri}^\prime)}$
        \\
        \Return $h = \sign(\hat{\pri}r - \hat{\cost}_0)$
    \end{algorithmic}
\end{algorithm}

\begin{algorithm}
    \caption{Stochastic optimization for non-negative Bregman divergence \citep{Kato2020}}
    \label{alg: nnBR}
    \begin{algorithmic}[1]
        \Require Positive and unlabeled dataset $(\setP, \setU)$, 
        \Ensure A trained model $\hat{r} : \mathbb{R}^d \to [0, \infty)$
        \While {no stopping criterion has been met}
            \State Create $N$ mini-batches $B_1, \ldots, B_N$
            \For {$i=1$ to $N$}
                \If {$\empexpectU{\mathfrak{F}(r(X))} - \alpha \empexpectP{\mathfrak{F}(r(X))} \geq 0$}
                    \State Set gradient: $\nabla_r \left( \empexpectP{-\di{f}(r(X))} + \empexpectU{f^*(r(X))} \right)$
                \Else
                    \State Set gradient: $\nabla_r \left( -\empexpectU{\mathfrak{F}(r(X))} + \alpha \empexpectP{\mathfrak{F}(r(X))} \right)$
                \EndIf
                \State Update $r$
            \EndFor
        \EndWhile
    \end{algorithmic}
\end{algorithm}

%% file: discussions.tex

In this section, we provide further theoretical analysis and compare the convergence rate of the proposed method to that of unbiased/non-negative PU classification.

\subsection{Selection of the Bregman generator function} \label{subsec: tightness}
Theorem \ref{thm: classification_shift} needs the assumption of strong convexity on the Bregman generator function $f$. We used a quadratic function in the proposed method; nevertheless there could be other choices of $f$. The following proposition shows that the tightest excess risk bound is achieved by a quadratic function.
\begin{proposition} \label{thm: generator_of_BR}
    Let $f$ be a strongly convex function and $\mu = \inf_{t \in [0, \infty)} \ddi{f}(t)$. Then the quadratic function $f_{\mathrm{S}}(t) = \mu t^2 / 2$ satisfies
    \begin{equation}
        \BR{f_\mathrm{S}}{r^*}{r} \leq \BR{f}{r^*}{r}.
    \end{equation}
\end{proposition}
Furthermore, Bregman divergence with respect to the quadratic function can be related to the excess classification risk with respect to the squared loss function. Let us denote the classification risk w.r.t.~the squared loss as
\begin{equation}
    R_\mathrm{sq}(g) = \expectm{X, Y}{\frac{1}{4}(Yg(X) - 1)^2},
\end{equation}
where $g: \mathbb{R}^d \to [-1, 1]$, and the optimal risk as
\begin{equation}
    R_\mathrm{sq}^* = \inf_g R_\mathrm{sq}(g) = R_\mathrm{sq}(2\eta - 1),
\end{equation}
where $\eta(x) = P(Y = +1 \mid X = x)$. Then, the Bregman divergence is decomposed into the excess risk w.r.t.~the squared loss and a superfluous term.
\begin{proposition} \label{thm: squared_loss}
    Let $g_r = 2 \min (\pri r, 1) - 1$ for any $r: \mathbb{R}^d \to [0, \infty)$. Then,
    \begin{equation}
        \frac{2 \pri^2}{\mu} \BR{f_\mathrm{S}}{r^*}{r} = R_\mathrm{sq}(g_r) - R_\mathrm{sq}^* + \chi_r \mathfrak{D}_r,
    \end{equation}
    where $\chi_r = (1 / \pri^2) \expectm{X \mid \pri r(X) > 1}{(\pri r(X) - 1)(\pri r(X) - 2 \eta(X) + 2)}$ and $\mathfrak{D}_r = P(\pi r(X) > 1)$.
\end{proposition}
If $r$ is bounded above by $1 / \pri$, the superfluous term is canceled and the Bregman divergence corresponds to the excess risk w.r.t.~the squared loss, up to the scaling factor.

\subsection{Excess risk bound for AUC} \label{subsec: auc}
Here we consider the relationship between AUC optimization and DRE. It is clear that the optimal density ratio $r^*$ is the optimal score function \citep{Menon2016br}, and as Theorem \ref{thm: classification_ratio}, we can obtain an excess AUC risk bound by the Bregman divergence of DRE as follows:
\begin{theorem} \label{thm: excess_auc}
    Let $f$ be a $\mu$-strongly convex function. Then, for any $r : \mathbb{R} \to [0, \infty)$, we have
    \begin{equation}
        R_\mathrm{AUC}(r) - R_\mathrm{AUC}^* \leq \frac{1}{1 - \pri} \sqrt{\frac{2}{\mu} \BR{f}{r^*}{r}}.
    \end{equation}
\end{theorem}
Theorem \ref{thm: excess_auc} implies that a better estimation of the density ratio in the sense of the Bregman divergence tends to result in a better score function in the sense of AUC.


\subsection{Excess risk bound with the estimated threshold}
Theorem \ref{thm: classification_shift} gives an excess risk bound with the optimal threshold. However, in practice, we need to use an estimated threshold. Here we also consider an excess risk bound for that case. Let $\theta$ be the true classification threshold for $h = \sign(r - \theta)$, defined as $\theta = \cost_0 / \pri$, and $\hat{\theta}$ be the empirical version of $\theta$, obtained from $\hat{\pri}(r)$ and $\hat{\pri}^\prime(r)$. Then, we have the following excess risk bound.
\begin{theorem} \label{thm: classification_threshold}
    Let $f$ be a $\mu$-strongly convex function and $\theta = \cost_0 / \pri$ where $\cost_0$ is defined in Theorem \ref{thm: classification_shift}. Then for $\hat{\theta} \in (0, 1)$ and $h_{\hat{\theta}} = \sign(r - \hat{\theta})$, we have
    \begin{equation}
        R_{\pris, \costs}(h_{\hat{\theta}}) - R_{\pris, \costs}^* \leq C \left( \left( 1 + \omega_{\hat{\theta}} \right) \sqrt{\frac{2}{\mu} \BR{f}{r^*}{r}} +  \abs{\hat{\theta} - \theta} \right),
    \end{equation}
    where $C$ is defined in Theorem \ref{thm: classification_shift} and $\omega_{\hat{\theta}}$ is a constant such that $0 \leq \omega_{\hat{\theta}} \leq 1$ and $\omega_{\theta} = 0$.
\end{theorem}
Theorem \ref{thm: classification_threshold} reduces to Theorem \ref{thm: classification_shift} when $\hat{\theta} = \theta$. We can also prove that the estimation error of the threshold decays at the linear order of the estimation error of the class-priors as follows:
\begin{proposition} \label{thm: threshold_error}
    Let $\hat{\pri}$, $\hat{\pri}^\prime$ be estimated class-priors and $\hat{\theta}$ be an estimated threshold by $\hat{\pri}$, $\hat{\pri}^\prime$. Then,
    \begin{equation}
        \abs{\hat{\theta} - \theta} \leq \mathcal{O}\left( \abs{\hat{\pri} - \pri} + \abs{\hat{\pri}^\prime - \pris} \right) \quad \mathrm{as} \ \abs{\hat{\pri} - \pri}, \abs{\hat{\pri}^\prime - \pris} \to 0.
    \end{equation}
\end{proposition}
Combining Corollary \ref{thm: classification_threshold} and Proposition \ref{thm: threshold_error}, we can see that the excess risk decays at the linear order of the estimation error of the class-priors.


\subsection{Convergence rate comparison to unbiased/non-negative PU classification} \label{subsec: comparison}
From the above results and theoretical analysis for non-negative Bregman divergence minimization provided by \citet{Kato2020}, we can derive the convergence rate for our proposed method. Let $\mathcal{H}$ be a hypothesis space of density ratio model $r : \mathbb{R}^d \to [0, \infty)$ and let us denote the minimizer of the empirical risk as $\hat{r} = \mathrm{argmin}_{r\in \mathcal{H}} \widetilde{\mathcal{L}}_f(r)$ where $\mathcal{L}_f$ and $\widetilde{\mathcal{L}}_f$ are defined in Section \ref{subsec: implementation}. Theorem 1 in \citet{Kato2020} states that if $f$ satisfies some appropriate conditions and the Rademacher complexity of $\mathcal{H}$ decays at $\mathcal{O}(1/\sqrt{n})$ w.r.t.~sample size $n$, for example, linear-in-parameter models with a bounded norm or neural networks with a bounded Frobenius norm \citep{Golowich2018,Lu2019}, the estimation error $\mathcal{L}(\hat{r}) - \inf_{r \in \mathcal{H}} \mathcal{L}(r)$ decays at $\mathcal{O}(1/\sqrt{\numP} + 1/\sqrt{\numU})$ with high probability. Applying this to Corollary \ref{thm: classification_threshold} and Proposition \ref{thm: threshold_error}, the following theorem is induced.
\begin{corollary} \label{thm: convergence}
    Let $f$ be a $\mu$-strongly convex function and satisfy Assumption 3 in \citet{Kato2020}. Then, for $\hat{h}_{\hat{\theta}} = \sign(\hat{r} - \hat{\theta})$,  with probability at least $1 - \delta$, we have
    \begin{equation}
        R_{\pris, \costs}(\hat{h}_{\hat{\theta}}) - R_{\pris, \costs}^* \leq A_\mathcal{H} + \mathcal{O} \left( \frac{1}{\numP^{1/4}} + \frac{1}{\numU^{1/4}} \right)  + \mathcal{O} \left( \abs{\hat{\pri} - \pri} + \abs{\hat{\pri}^\prime - \pris} \right),
    \end{equation}
   where $A_\mathcal{H} = C \sqrt{\frac{2}{\mu} \left( \inf_{r \in \mathcal{H}}\mathcal{L}_f(r) - \mathcal{L}_f(r^*) \right) }$ with the constant $C$ defined in Theorem \ref{thm: classification_shift} and $\delta$ gives a scale factor $\sqrt{\log (1 / \delta)}$ to the term of $\numP$ and $\numU$.
\end{corollary}
\par
For comparison, we consider the convergence of the excess risk based on the theoretical analysis of unbiased/non-negative PU classification provided by \citet{Kiryo2017} and the properties of \emph{classification calibrated} loss functions provided by \citet{Bartlett2006} and \citet{Scott2012}. Let $\mathcal{G}$ be a hypothesis space of decision function $g : \mathbb{R}^d \to \mathbb{R}$ and let us denote the minimizer of the empirical risk as $\hat{g} = \inf_{g \in \mathcal{G}}\widetilde{R}_\ell(g)$ where $R_\ell$ and $\widetilde{R}_\ell$ are defined in Section \ref{sec: relatedwork}. Assume that the loss function $\ell$ satisfies some appropriate conditions and if the Rademacher complexity of $\mathcal{G}$ decays at $\mathcal{O}(1/\sqrt{n})$, the estimation error $R_\ell(\hat{g}) - \inf_{g \in \mathcal{G}} R_\ell(g)$ decays at $\mathcal{O}(1/\sqrt{\numP} + 1/\sqrt{\numU})$ with high probability. In addition, if $\ell$ is classification calibrated \citep{Bartlett2006,Scott2012}, there exists a strictly increasing function $\psi$ and the excess risk w.r.t.~the zero-one loss is bounded above by the surrogate excess risk. That is, with probability at least $1 - \delta$, we have
\begin{equation}
    R_{\pri, \cost}(\hat{g}) - R_{\pri, \cost}^* \leq \psi^{-1} \left( A_\mathcal{G} + \mathcal{O} \left( \frac{1}{\sqrt{\numP}} + \frac{1}{\sqrt{\numU}} \right) \right),
\end{equation}
where $A_\mathcal{G} = \inf_{g \in \mathcal{G}} R_\ell(g) - R_\ell^*$, and $\delta$ gives a scale factor $\sqrt{\log (1 / \delta)}$ to the term of $\numP$ and $\numU$.
\par 
For specific loss functions such as the hinge loss or sigmoid loss, $\psi$ is the identity function \citep{Bartlett2006,Steinwart2007}, hence the convergence rate of unbiased/non-negative PU classification would be faster than that of the density ratio estimation approach for PU classification (DRPU). This result is intuitively reasonable, because a method solving a specific problem tends to have better performance than a method solving more general problems \citep{Vapnik1995}. That is, the hinge loss and sigmoid loss are not \emph{proper losses} in the context of class-posterior probability estimation \citep{Buja2005,Reid2009}, and risk minimization with respect to these losses allows one to bypass the estimation of the posterior probability and obtain a classifier directly, while DRE does not.

%% file: experiments.tex

In this section, we report our experimental results. All the experiments were done with \emph{PyTorch} \citep{Paszke2019}. \footnote{We downloaded the source-codes of nnPU from \url{https://github.com/kiryor/nnPUlearning}, VPU from \url{https://github.com/HC-Feynman/vpu}, and KM2 from \url{http://web.eecs.umich.edu/~cscott/code.html}. Our implementation is available at \url{https://github.com/csnakajima/pu-learning}.}

\subsection{Test with synthetic data}
We conducted experiments with synthetic data to confirm the effectiveness of the proposed method via numerical visualization. Firstly, we define $p_+(x) = \mathcal{N}(+1, 1)$ and $p_-(x) = \mathcal{N}(-1, 1)$ where $\mathcal{N}(\mu, \sigma^2)$ denotes the univariate Gaussian distribution with mean $\mu$ and variance $\sigma^2$, and $p(x) = \pri p_+(x) + (1 - \pri) p_-(x)$. We generated samples from $p_+(x)$ and $p(x)$ with $\pri = 0.4$ for training data, and from $p(x)$ with $\pris = 0.6$ for test data.
\par
The training dataset contained 200 positively labeled samples and 1000 unlabeled samples, and the validation dataset contained 100 positively labeled samples and 500 unlabeled samples. The test dataset consisted of 1000 samples. As a parametric model, linear-in-parameter model with Gaussian basis functions $\varphi(x) = \exp(-(x - x_i)^2 / 2)$, where $\{x_1, \ldots , x_{\numU}\} = \setU$, was used. Adam with default momentum parameters $\beta_1 = 0.5$, $\beta_2 = 0.999$ and $\ell_2$ regularization parameter $0.1$ was used as an optimizer. Training was performed for 200 epochs with the batch size $200$ and the learning rate $2e-5$.
\par
\begin{figure}[ht]
    \centering
    \includegraphics[width=1.0\linewidth]{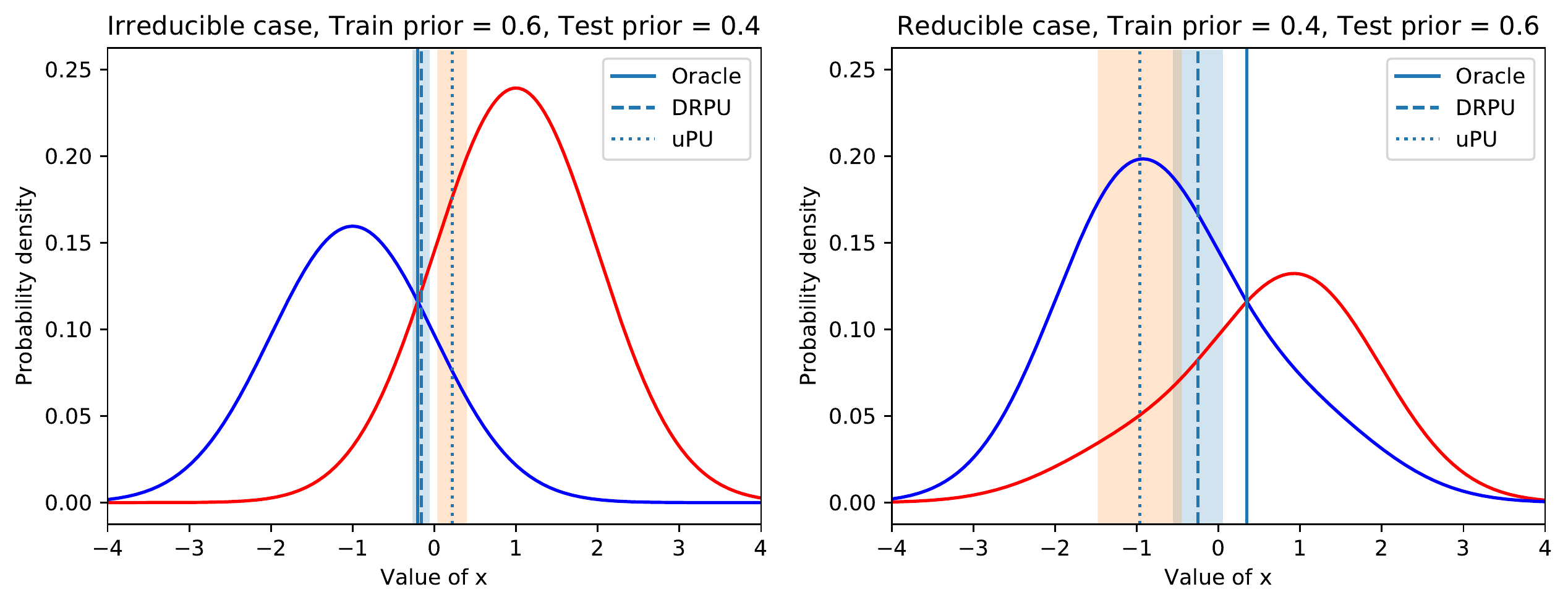}
    \caption{Visualized classification boundaries of uPU and DRPU, averaged over 10 trials. Each of the vertical lines are the boundaries and the colored areas are the standard deviations. ``Oracle'' is the optimal classification boundary. The red curve means the probability density $p_+$ scaled by $\pris$, and the blue curve means $p_-$ scaled by $1 - \pris$. The upper graph corresponds to the case where irreduciblity assumption holds, while the lower one does not.}
    \label{fig: synthetic}
\end{figure}

 
We did experiments with unbiased PU learning (uPU) \citep{DuPlessis2014pu,DuPlessis2015} with the logistic loss and our proposed method (DRPU) with LSIF. In uPU, the class-prior of the training data was estimated by KM2 \citep{Ramaswamy2016}, and for DRPU, the test unlabeled dataset was used as an unlabeled dataset to estimate the test prior $\pris$. The left-hand side of Figure \ref{fig: synthetic} shows the obtained classification boundaries, and the boundary of DRPU was closer to the optimal one than that of uPU.
\par
Secondly, we tested the case where the irreducibility assumption does not hold. Let $p_+(x) = 0.8 \mathcal{N}(+1, 1) + 0.2 \mathcal{N}(-1, 1)$, $p_-(x) = 0.2 \mathcal{N}(+1, 1) + 0.8 \mathcal{N}(-1, 1)$, and set the training prior $\pri = 0.6$, the test prior $\pris = 0.4$. The result is illustrated in the right-hand side of figure \ref{fig: synthetic}. Class-prior estimation by KM2 was inaccurate since the irreducibility assumption did not hold, and then uPU led to a large error. DRPU also gave inaccurate estimations of the class-priors, but they did not affect the training step, so the influence of the estimation error was relatively mitigated.

\subsection{Test with benchmark data}

\begin{table}[ht]
    \centering
    \caption{The means and standard deviations of the classification accuracy in percent on benchmark datasets over 10 trials. ``Train'' and ``Test'' denote the class-priors of the training and test data respectively. ``Avg'' is the averaged accuracy of the four results with different priors. The best results with respect to the one-sided {\it t}-test at the significance level 0.05 are highlighted in boldface (for the ``Avg'' case, just picking the highest one).}
    \setlength{\tabcolsep}{1.0pt}
    \begin{tabular*}{\linewidth}{@{\extracolsep{\fill}}ccccccc}
        \toprule
        Dataset & Train & Test & nnPU & PUa & VPU & DRPU  \\
        \midrule
        \multirow{5}{*}{MNIST} & \multirow{5}{*}{$0.5$} & $0.2$ & \avg{92.98}{1.72} & \avg{93.11}{1.51} & \avgbold{96.21}{0.20} & \avg{95.78}{0.54} \\
        & & $0.4$ & \avg{93.76}{1.02} & \avg{93.70}{0.94} & \avgbold{94.93}{0.51} & \avgbold{94.85}{0.58} \\
        & & $0.6$ & \avgbold{94.52}{0.31} & \avgbold{94.79}{0.48} & \avg{93.71}{0.89} & \avgbold{94.67}{0.46} \\
        & & $0.8$ & \avg{95.23}{0.76} & \avg{95.28}{0.90} & \avg{92.43}{1.39} & \avgbold{95.91}{0.48} \\
        & & Avg. & $94.12$ & $94.22$ & $94.32$ & $\mathbf{95.30}$ \\
        \midrule
        \multirow{5}{*}{
            \begin{tabular}{c}
                Fashion- \\
                MNIST
            \end{tabular}
        } & \multirow{5}{*}{$0.6$} & $0.2$ & \avgbold{91.67}{0.97} & \avgbold{91.05}{0.61} & \avg{89.22}{1.13} & \avgbold{91.59}{0.57} \\
        & & $0.4$ & \avg{88.95}{1.22} & \avg{87.89}{1.10} & \avgbold{90.11}{0.51} & \avgbold{90.42}{0.92} \\
        & & $0.6$ & \avg{86.26}{1.50} & \avg{86.07}{1.35} & \avgbold{90.80}{0.48} & \avgbold{90.46}{0.72} \\
        & & $0.8$ & \avg{83.26}{2.29} & \avg{84.39}{2.27} & \avg{91.70}{0.98} & \avgbold{92.97}{0.58} \\
        & & Avg. & $87.54$ & $87.35$ & $90.46$ & $\mathbf{91.36}$ \\
        \midrule
        \multirow{5}{*}{
            \begin{tabular}{c}
                Kuzushiji- \\
                MNIST
            \end{tabular}
        } & \multirow{5}{*}{0.4} & 0.2 & \avg{81.88}{2.52} & \avg{82.81}{2.99} & \avg{85.78}{2.81} & \avgbold{91.41}{0.56} \\
        & & 0.4 & \avg{85.18}{1.64} & \avg{85.11}{2.12} & \avg{85.93}{1.77} & \avgbold{88.62}{0.52} \\
        & & 0.6 & \avgbold{88.35}{0.95} & \avgbold{87.87}{0.94} & \avg{86.11}{1.36} & \avgbold{88.22}{0.67} \\
        & & 0.8 & \avgbold{91.28}{0.53} & \avgbold{90.67}{1.73} & \avg{86.10}{2.09} & \avgbold{91.32}{0.62} \\
        & & Avg. & $86.67$ & $86.62$ & $85.98$ & $\mathbf{89.89}$ \\
        \midrule
        \multirow{5}{*}{CIFAR-10} & \multirow{5}{*}{0.4} & 0.2 & \avg{80.32}{1.55} & \avg{80.33}{2.34} & \avgbold{91.94}{0.67} & \avgbold{91.59}{0.37} \\
        & & 0.4 & \avg{84.38}{1.15} & \avg{84.03}{1.59} & \avgbold{89.00}{1.21} & \avgbold{88.67}{0.56} \\
        & & 0.6 & \avg{88.16}{0.54} & \avg{88.05}{0.95} & \avg{85.99}{2.45} & \avgbold{89.22}{0.56} \\
        & & 0.8 & \avgbold{92.28}{0.51} & \avgbold{92.21}{0.41} & \avg{82.92}{3.92} & \avgbold{92.45}{0.35} \\
        & & Avg. & $86.28$ & $86.16$ & $87.46$ & $\mathbf{90.48}$ \\
        \bottomrule
    \end{tabular*}
    \label{tab: accuracy}
\end{table}

\begin{figure*}[ht]
    \centering
    \includegraphics[width=1.0\linewidth]{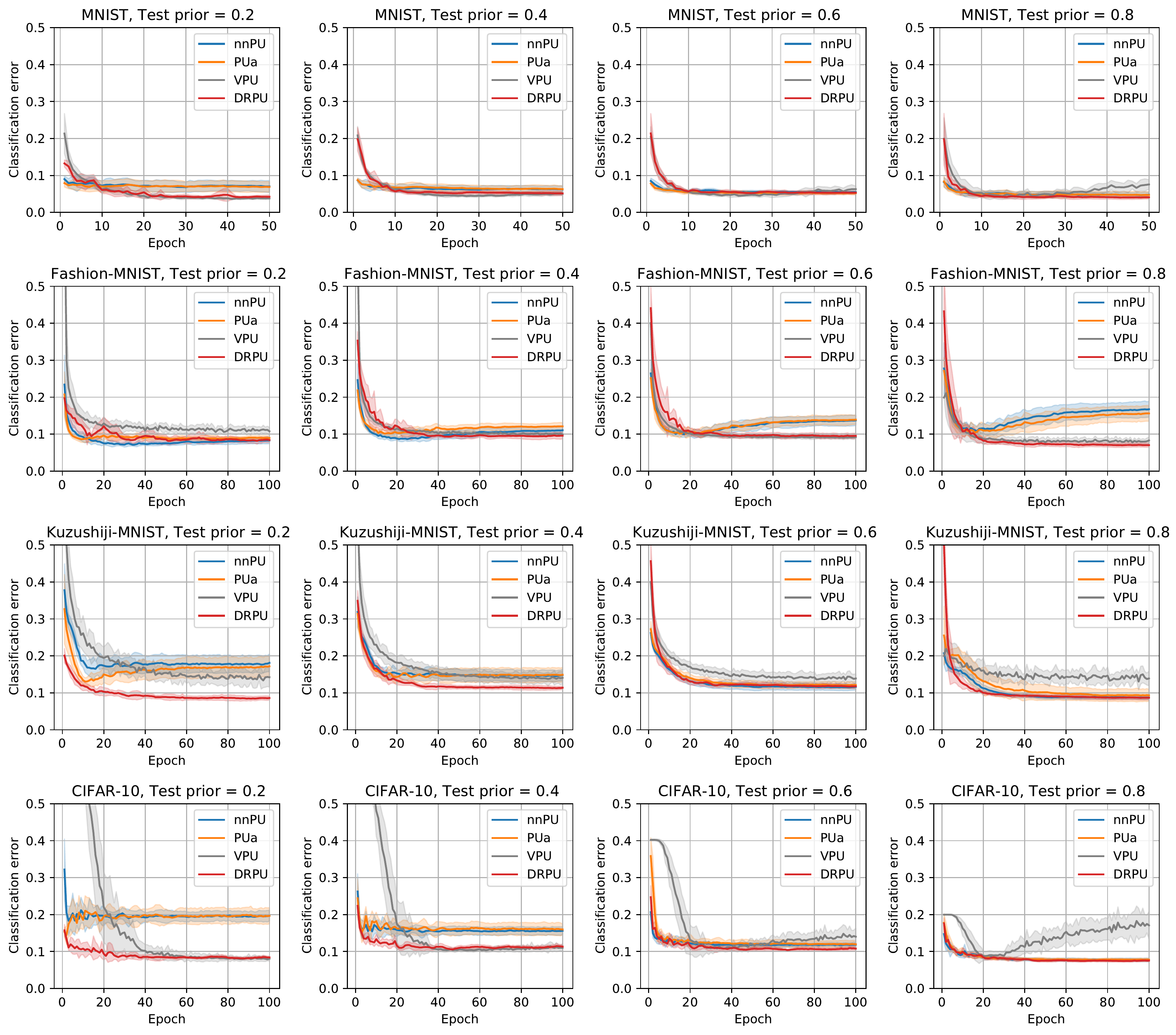}
    \caption{The means and standard deviations of the classification error as functions of the training epoch.}
    \label{fig: test}
\end{figure*}

We also measured the performances of nnPU \citep{Kiryo2017}, PUa \citep{Charoenphakdee2019}, VPU \citep{Chen2020}, and DRPU (the proposed method) on MNIST \citep{LeCun1998}, Fashion-MNIST \citep{xiao2017fashionmnist}, Kuzushiji-MNIST \citep{Lamb2018kuzushiji}, and CIFAR-10 \citep{Alex2012cifar}. 
Here we summarize the descriptions of the datasets and the training settings.

\begin{itemize}
    \item MNIST \citep{LeCun1998} is a gray-scale 28 $\times$ 28 image dataset of handwritten digits from 0 to 9, which contains 60000 training samples and 10000 test samples. Since it has 10 classes, we treated the even digits as the positive class and the odd digits as the negative class respectively. We prepared 2500 positively labeled (P) samples and 50000 unlabeled (U) samples as the training data, and 500 P samples and 10000 U samples as the validation data. The test dataset was made up of 5000 samples for each of the test distributions with different class-priors respectively. As a parametric model, 5-layer MLP : 784-300-300-300-1 with ReLU activation was used, and trained by Adam with default momentum parameters $\beta_1 = 0.9$, $\beta_2 = 0.999$ and $\ell_2$ regularization parameter $5e-3$. Training was performed for 50 epochs with the batch size $500$. The learning rate was set to $1e-4$ for nnPU/PUa and $2e-5$ for VPU/DRPU, which would be halved for every 20 epochs. In VPU, we set hyperparameters for Mixup as $\alpha=0.3$ and $\lambda=2.0$. In DRPU, we set a hyperparameter for non-negative correction as $\alpha=0.475$. 
    \item Fashion-MNIST \citep{xiao2017fashionmnist} is a gray-scale 28 $\times$ 28 image dataset of 10 kinds of fashion items, which contains 60000 training samples and 10000 test samples. We treated `Pullover', `Dress', `Coat', `Sandal', `Bag', and `Ankle boot' as the positive class, and `T-shirt', `Trouser', `Shirt', and `Sneaker' as the negative class respectively. We prepared 2500 P samples and 50000 U samples as training data, and 500 P samples and 10000 U samples for validation data. The test dataset was made up of 5000 samples for each of the test distributions with different class-priors respectively. As a parametric model, we used LeNet \citep{LeCun1998} -based CNN : (1 $\times$ 32 $\times$ 32) - C(6, 5 $\times$ 5, pad=2) - MP(2) - C(16, 5 $\times$ 5, pad=2) - MP(2) - C(120, 5 $\times$ 5) - 120 - 84 - 1, where C($c$, $h \times w$, pad=$p$) means $c$ channels of $h \times w$ convolutions with zero-padding $p$ (abbreviated if $p=0$) followed by activation function (ReLU), and MP($k$) means $k \times k$  max pooling. Batch normalization was applied after the first fully-connected layer. The model was trained by Adam, with the same settings as the case of MNIST. Training was performed for 100 epochs with the batch size $500$ and the learning rate $2e-5$, which would be halved for every 20 epochs. In VPU, we set hyperparameters for Mixup as $\alpha=0.3$ and $\lambda=0.5$. In DRPU, we set a hyperparameter for non-negative correction as $\alpha=0.6$.
    \item Kuzushiji-MNIST \citep{Lamb2018kuzushiji} is a gray-scale 28 $\times$ 28 image dataset of 10 kinds of cursive Japanese characters, which contains 60000 training samples and 10000 test samples. We treated `o', `ki', `re', `wo' as the positive class, and `su', `tsu', `na', `ha', `ma', `ya' as the negative class respectively. We prepared 2500 P samples and 50000 U samples as the training data, and 500 P samples and 10000 U samples as the validation data. The test dataset was made up of 5000 samples for each of the test distributions with different class-priors respectively. The model and optimization settings were the same as the cases of Fashion-MNIST. In VPU, we set hyperparameters for Mixup as $\alpha=0.3$ and $\lambda=0.5$. In DRPU, we set a hyperparameter for non-negative correction as $\alpha=0.375$.
    \item CIFAR-10 \citep{Alex2012cifar} is a colored 32 $\times$ 32 image dataset, which contains 50000 training samples and 10000 test samples. We treated `airplane', `automobile', `ship', and `truck' as the positive class, and `bird', `cat', `deer', `dog', `frog', and `horse' as the negative class respectively. We prepared 2500 P samples and 45000 U samples as the training data, and 500 P samples and 5000 U samples as the validation data. The test dataset was made up of 5000 samples for each the test distributions with different class-priors respectively. As a parametric model, we used the CNN introduced in \citet{Springenberg2014} : (3 $\times$ 32 $\times$ 32) - C(96, 5 $\times$ 5, pad=2) - MP(2 $\times$ 2) - C(96, 5 $\times$ 5, pad=2) - MP(2 $\times$ 2) - C(192, 5 $\times$ 5, pad=2) - C(192, 5 $\times$ 3, pad=1) - C(192, 1 $\times$ 1) - C(10, 1 $\times$ 1) with ReLU activation. Batch normalization was applied after the max pooling layers and the third, fourth, fifth convolution layers. The model was trained by Adam, with the same settings as the case of MNIST. Training was performed for 100 epochs with the batch size $500$ and the learning rate $1e-5$, which would be halved for every 20 epochs. In VPU, we set hyperparameters for Mixup as $\alpha=0.3$ and $\lambda=4.0$. In DRPU, we set a hyperparameter for non-negative correction as $\alpha=0.425$.
\end{itemize}

\begin{table}[ht]
    \centering
    \caption{The means and standard deviations of the AUC in percent on benchmark datasets over 10 trials. The best results with respect to the one-sided \emph{t}-test at the significance level 0.05 are highlighted in boldface.}
    \begin{tabular}{ccccc}
        \toprule
        Dataset & nnPU & PUa & VPU & DRPU  \\
        \midrule
        MNIST & \avgtwo{0.9855}{0.0022} & \avgtwo{0.9861}{0.0024} & \avgboldtwo{0.9902}{0.0010} & \avgtwo{0.9815}{0.0023} \\
        \midrule
        F-MNIST & \avgtwo{0.9371}{0.0129} & \avgtwo{0.9359}{0.0065} & \avgboldtwo{0.9656}{0.0032} & \avgtwo{0.9607}{0.0055} \\
        \midrule
        K-MNIST & \avgboldtwo{0.9505}{0.0050} & \avgboldtwo{0.9485}{0.0044} & \avgtwo{0.9347}{0.0129} & \avgboldtwo{0.9507}{0.0033} \\
        \midrule
        CIFAR-10 & \avgtwo{0.9509}{0.0062} & \avgtwo{0.9512}{0.0071} & \avgboldtwo{0.9594}{0.0024} & \avgboldtwo{0.9577}{0.0028} \\
        \bottomrule
    \end{tabular}
    \label{tab: auc}
\end{table}

In nnPU, the class-prior of the training data was estimated by KM2 \citep{Ramaswamy2016}. In PUa, we estimated both the training and test priors by KM2, then performed cost-sensitive non-negative PU classification \citep{Charoenphakdee2019}. Note that in this setting, PUa needs the unlabeled test dataset at the training-time to estimate the test prior by KM2, while DRPU needs it at only the test-time. Moreover, PUa needs to train a model for each time the test prior changes. In nnPU and PUa, the sigmoid loss was used as a loss function. 
\par 
Table \ref{tab: accuracy} shows the results of the experiments. nnPU and PUa unintentionally achieved high accuracy in some cases because of estimation errors of the class-priors, while they had poor results in the other cases. VPU achieved good results in several cases where the scale of class-prior shift was small, but it was not adapted to large class-prior shift. DRPU outperformed the other methods in almost all cases, and was the most robust to the test-time class-prior shift. Figure \ref{fig: test} gives the classification errors in the experiments. For example, in the case of Fashion-MNIST with $\pris = 0.8$, nnPU and PUa suffered from overfitting due to the estimation error of the training class-prior. Also, as seen in the case of Kuzushiji-MNIST with $\pris = 0.2$, DRPU gave a better result than the other methods, and was the most stable, i.e., it had the smallest variance.
\par
In addition, Table \ref{tab: auc} reports the computed AUC values on the experiments for each of the methods. The results were picked from $\pris = 0.6$ case. DRPU had a bit worse results than VPU on MNIST and Fashion-MNIST, while it performed well on Kuzushiji-MNIST and CIFAR-10. Table \ref{tab: abs_prior} summarizes the absolute error of the class-prior estimation by KM2 and our method described in Section \ref{subsec: prior}. For KM2, we used 2000 positive samples from the training dataset and 2000 unlabeled samples from the test dataset. The inputs were transformed into 50 dimensions by PCA \citep{Jolliffe2016}. For our method, we used 500 positive samples from the validation dataset and 5000 unlabeled samples from the test dataset. It is observed that our class-prior estimation method outperformed KM2 in almost all cases.

\begin{table}[ht]
    \centering
    \caption{The means and standard deviations of the absolute error of the class-prior estimation in percent on benchmark datasets over 10 trials. The best results with respect to one-sided \emph{t}-test at the significance level 0.05 are highlighted in boldface.}
    \begin{tabular}{cccc}
        \toprule
        Dataset & Prior & KM2 & Ours \\
        \midrule
        \multirow{4}{*}{MNIST} & 0.2 & \avg{0.0679}{0.0147} & \avgbold{0.0101}{0.0128} \\
        & 0.4 & \avg{0.0410}{0.0169} & \avgbold{0.0279}{0.0265} \\
        & 0.6 & \avgbold{0.0209}{0.0110} & \avgbold{0.0361}{0.0372} \\
        & 0.8 & \avg{0.2885}{0.0560} & \avgbold{0.0485}{0.0429} \\
        \midrule
        \multirow{4}{*}{F-MNIST} & 0.2 & \avgbold{0.0133}{0.0105} & \avg{0.0230}{0.0120} \\
        & 0.4 & \avgbold{0.0170}{0.0151} & \avgbold{0.0181}{0.0115} \\
        & 0.6 & \avg{0.0843}{0.0297} & \avgbold{0.0183}{0.0148} \\
        & 0.8 & \avg{0.2534}{0.0340} & \avgbold{0.0249}{0.0231} \\
        \midrule
        \multirow{4}{*}{K-MNIST} & 0.2 & \avg{0.0326}{0.0155} & \avgbold{0.0220}{0.0130} \\
        & 0.4 & \avg{0.1029}{0.0201} & \avgbold{0.0704}{0.0211} \\
        & 0.6 & \avg{0.2685}{0.0246} & \avgbold{0.1088}{0.0310} \\
        & 0.8 & \avg{0.4869}{0.0361} & \avgbold{0.1496}{0.0394} \\
        \midrule
        \multirow{4}{*}{CIFAR-10} & 0.2 & \avg{0.1880}{0.0097} & \avgbold{0.0151}{0.0112} \\
        & 0.4 & \avg{0.1399}{0.0163} & \avgbold{0.0189}{0.0142} \\
        & 0.6 & \avg{0.0738}{0.0167} & \avgbold{0.0305}{0.0184} \\
        & 0.8 & \avg{0.1242}{0.0784} & \avgbold{0.0397}{0.0322} \\
        \bottomrule
    \end{tabular}
    \label{tab: abs_prior}
\end{table}
\renewcommand{\arraystretch}{1}


\subsection{Comparisons under different numbers of labeled samples}

\begin{figure}[ht]
    \centering
    \includegraphics[width=1.0\linewidth]{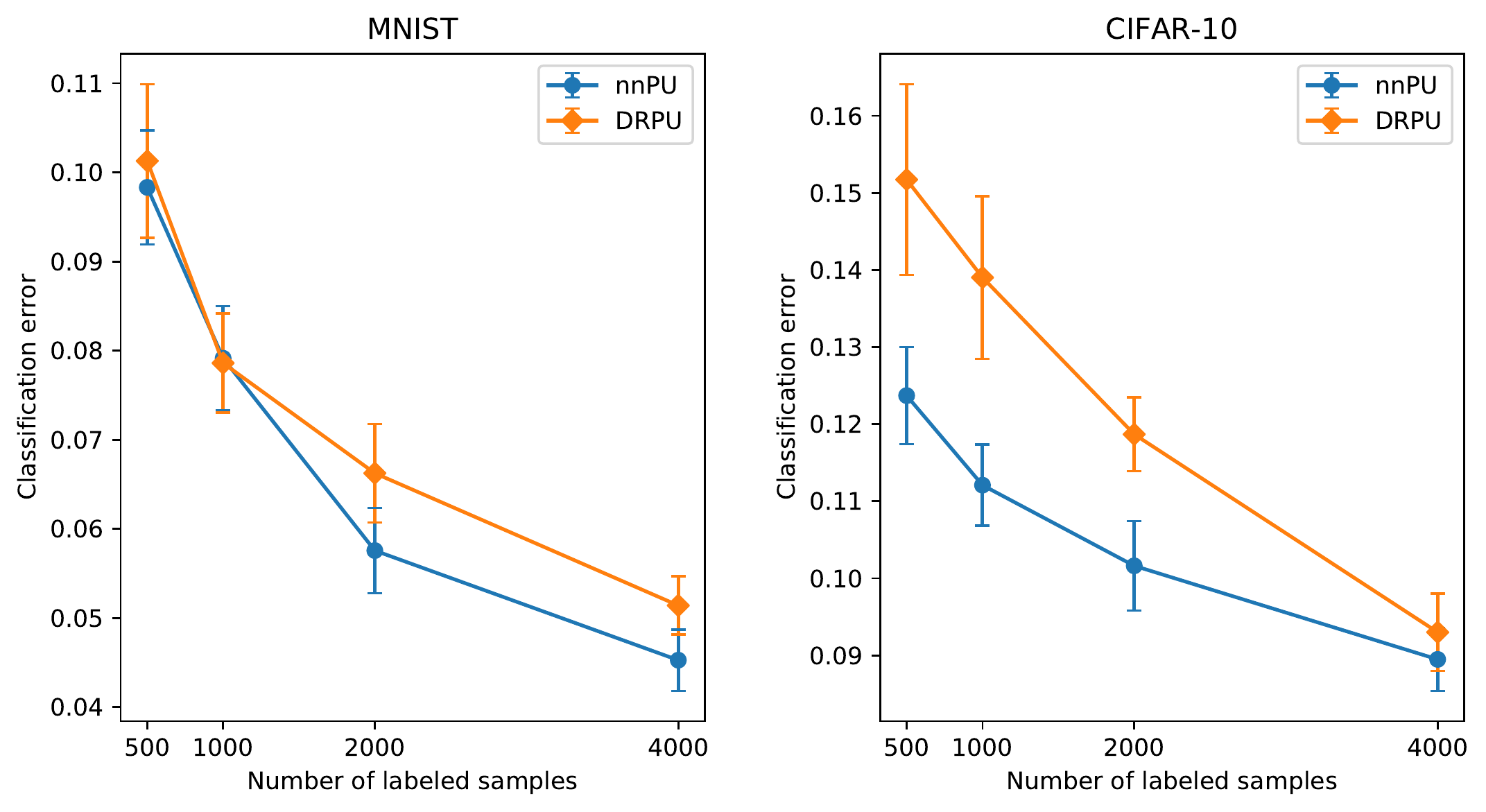}
    \caption{Classification errors of nnPU and DRPU on MNIST and CIFAR-10, averaged over 10 trials for each settings of the number of labeled samples. The vertical bars at each of the points refer the standard deviations.}
    \label{fig: comparison}
\end{figure}

To numerically verify the theoretical insight provided in Section \ref{subsec: comparison}, we compared nnPU and DRPU with different sizes of the positively labeled dataset. In this experiment, we assumed that the true class-prior $\pri$ was known and no class-prior shift would occur. We performed nnPU and DRPU on MNIST and CIFAR-10, with $\numP \in \{500, 1000, 2000, 4000\}$. Note that we skipped the class-prior estimation step of DRPU because the class-priors were given. Figure \ref{fig: comparison} shows the results of the experiments. On MNIST, the performance of DRPU was comparable to that of nnPU when $\numP$ was small, yet it got outperformed under larger $\numP$. On CIFAR-10, unlike the MNIST case, the difference in the classification error was larger when $\numP$ was smaller. As a whole, nnPU stably outperformed DRPU, and this experimental result supports the theoretical discussion in Section \ref{subsec: comparison}.

%% file: conclusions.tex


In this paper, we investigated positive-unlabeled (PU) classification from a perspective of density ratio estimation, and proposed a novel PU classification method based on density ratio estimation. The proposed method does not require the class-priors in the training phase, and it can cope with class-prior shift in the test phase. We provided theoretical analysis for the proposed method, and demonstrated its effectiveness in the experiments. Extending our work to other weakly-supervised learning problems \citep{Lu2018uu,Bao18su} is a promising future work.

%% file: proofs.tex

\subsection{Proof of Theorem \ref{thm: classification_ratio}}
From Lemma 1 of \citet{Scott2012}, we have
\begin{align*}
    R_{\pri, \cost}(g) - R_{\pri, \cost}^* = \expectm{X}{\indi{ \sign(g(X)) \neq \sign(\eta(X) - \cost) } \abs{\eta(X) - \cost}},
\end{align*}
where $\eta(x) = p(Y=+1 \mid X = x)$. Then, for $h_\cost = \pri r - \cost$,
\begin{align*}
    R_{\pri, \cost}(h_\cost) - R_{\pri, \cost}^* &= \expectm{X}{\indi{ (\pri r(X) - \cost)(\eta(X) - \cost) < 0 } \abs{\eta(X) - \cost}} \\
    &= \expectm{X}{\indi{ \pri r(X) < \cost < \eta(X) } \abs{\eta(X) - \cost}}  \\
    &\quad + \expectm{X}{\indi{ \eta(X) < \cost < \pri r(X) } \abs{\eta(X) - \cost}} \\
    &\leq \expectm{X}{\indi{ \pri r(X) < \eta(X) } \abs{\eta(X) - \pri r(X)}} \\
    &\quad + \expectm{X}{\indi{ \eta(X) < \pri r(X) } \abs{\eta(X) - \pri r(X)}} \\
    &= \expectm{X}{\abs{\eta(X) - \pri r(X)}} \\
    &= \pri \expectm{X}{\abs{r^*(X) - r(X)}} \\
    &\leq \pri \sqrt{\expectm{X}{(r^*(X) - r(X))^2}} \\
    &= \pri \sqrt{\frac{2}{\mu} \expectm{X}{\frac{\mu}{2}(r^*(X) - r(X))^2}} \\
    &\leq \pri \sqrt{\frac{2}{\mu} \expectm{X}{f(r^*(X)) - f(r(X)) - \di{f}(r(X))(r^*(X) - r(X))}} \\
    &= \pri \sqrt{\frac{2}{\mu} \BR{f}{r^*}{r}},
\end{align*}
where the second inequality is Jensen's, and the third inequality comes from the definition of strong convexity.
\hfill $\qedsymbol$


\subsection{Proof of Theorem \ref{thm: classification_shift}}
Same as Theorem 1 of \citet{Charoenphakdee2019}, we normalize coefficients of $R_{\pri, \cost}$ and $R_{\pris, \costs}$ and determine $\cost$ to satisfy
\begin{align*} 
    \frac{R_{\pri, \cost}(g)}{(1 - \cost)\pri + \cost(1 - \pri)} = \frac{R_{\pris, \costs}(g)}{(1 - \costs)\pris + \costs(1 - \pris)}.
\end{align*}
Compare the coefficient of the $\expectP{\cdot}$,
\begin{align*}
    \frac{(1 - \cost)\pri}{(1 - \cost)\pri + \cost(1 - \pri)} = \frac{(1 - \costs)\pris}{(1 - \costs)\pris + \costs(1 - \pris)}.
\end{align*}
Solve this equation with respect to $\cost$ and denote it as $\cost_0$,
\begin{align*}
    \cost_0 = \frac{\costs \pri (1 - \pris)}{(1 - \costs)(1 - \pri)\pris + \costs \pri (1 - \pris)}.
\end{align*}
Therefore, we obtain
\begin{align*}
    R_{\pris, \costs}(h_{\cost_0}) - R_{\pris, \costs}^* &= \frac{(1 - \costs)\pris + \costs(1 - \pris)}{(1 - \cost_0)\pri + \cost_0(1 - \pri)} \left( R_{\pri, \cost_0}(h_{\cost_0}) - R_{\pri, \cost_0}^* \right) \\
    &\leq \pri \frac{\costs + \pris - 2\costs \pris}{\cost_0 + \pri - 2\cost_0 \pri} \sqrt{\frac{2}{\mu} \BR{f}{r^*}{r}},
\end{align*}
where $h_{\cost_0} = \sign(\pri r - \cost_0)$ and the inequality is from Theorem \ref{thm: classification_ratio}.
\hfill $\qedsymbol$


\subsection{Proof of Theorem \ref{thm: priorestimator}}
We separate $\abs{\hat{\pri} - \pri}$ as follows
\begin{align} \label{eq: decomposition}
    \abs{\hat{\pri} - \pri} \leq \abs{\hat{\pri} - \inf_{h \in \mathcal{H}_r} \frac{P(h)}{P_+(h)}} + \abs{\inf_{h \in \mathcal{H}_r} \frac{P(h)}{P_+(h)} - \pri}
\end{align}
The first term of Eq. (\ref{eq: decomposition}) is upper bounded by the uniform bound
\begin{align*}
    \abs{\hat{\pri} - \inf_{h \in \mathcal{H}_r} \frac{P(h)}{P_+(h)}} &= \abs{\inf_{h \in \mathcal{H}_r} \frac{\widehat{P}(h)}{\widehat{P}_+(h)} - \inf_{h \in \mathcal{H}_r} \frac{P(h)}{P_+(h)}} \\
    &\leq \max \left( \frac{\widehat{P}(h_1)}{\widehat{P}_+(h_1)} - \frac{P(h_1)}{P_+(h_1)}, \frac{P(h_2)}{P_+(h_2)} - \frac{\widehat{P}(h_2)}{\widehat{P}_+(h_2)} \right) \\
    &\leq \sup_{h \in \mathcal{H}_r} \abs{\frac{\widehat{P}(h)}{\widehat{P}_+(h)} - \frac{P(h)}{P_+(h)}}
\end{align*}
where $h_1 = \mathrm{argmin}_{h \in \mathcal{H}_r} \frac{P(h)}{P_+(h)}$, $h_2 = \mathrm{argmin}_{h \in \mathcal{H}_r} \frac{\widehat{P}(h)}{\widehat{P}_+(h)}$.
\par
From McDiarmid's inequality, with probability at least $1 - \delta$, we have
\begin{align*}
    \sup_{h \in \mathcal{H}_r} \abs{\widehat{P}(h) - P(h)} &\leq \Radem{\numU}{p}(\mathcal{H}_r) + \sqrt{\frac{\log \frac{2}{\delta}}{2\numU}} \\
    &\leq \sqrt{\frac{4\log(e\numU / 2)}{\numU}} + \sqrt{\frac{\log \frac{2}{\delta}}{2\numU}} \\
    &= \varepsilon(\numU, \delta)
\end{align*}
for any $0 < \delta < 1$. We used $\Radem{n}{p}(\mathcal{H}) \leq \sqrt{\frac{2d\log(en/d)}{n}}$ where $d$ is VC-dimension of $\mathcal{H}$, and $\mathrm{VCdim}(\mathcal{H}_r) \leq 2$ for a fixed $r$.
The same discussion holds for $\widehat{P}_+(h)$, with probability at least $1 - \delta$,
\begin{align*}
    \sup_{h \in \mathcal{H}_r} \abs{\widehat{P}_+(h) - P_+(h)} \leq \varepsilon(\numP, \delta)
\end{align*}
Thus, with probability at least $(1 - \delta_\rmP)(1 - \delta_\rmU)$, we have
\begin{align*}
    \frac{\widehat{P}(h) - \varepsilon(\numU, \delta_\rmU)}{\widehat{P}_+(h) + \varepsilon(\numP, \delta_\rmP)} \leq \frac{P(h)}{P_+(h)} \leq \frac{\widehat{P}(h) + \varepsilon(\numU, \delta_\rmU)}{\widehat{P}_+(h) - \varepsilon(\numP, \delta_\rmP)}
\end{align*}
For the left hand side,
\begin{align*}
    \frac{\widehat{P}(h) - \varepsilon(\numU, \delta_\rmU)}{\widehat{P}_+(h) + \varepsilon(\numP, \delta_\rmP)} &= \left( \frac{\widehat{P}(h)}{\widehat{P}_+(h)} - \frac{\varepsilon(\numU, \delta_\rmU)}{\widehat{P}_+(h)} \right) \sum_{i = 0}^\infty \left( -\frac{\varepsilon(\numP, \delta_\rmP)}{\widehat{P}_+(h)} \right) ^ i \\
    &= \frac{\widehat{P}(h)}{\widehat{P}_+(h)} - \frac{\varepsilon(\numU, \delta_\rmU)}{\widehat{P}_+(h)} \\
    &\ \quad - \left( \frac{\widehat{P}(h)}{\widehat{P}_+(h)} - \frac{\varepsilon(\numU, \delta_\rmU)}{\widehat{P}_+(h)} \right) \frac{\varepsilon(\numP, \delta_\rmP)}{\widehat{P}_+(h)} \sum_{i=1}^\infty \left( -\frac{\varepsilon(\numP, \delta_\rmP)}{\widehat{P}_+(h)} \right)^{i-1} \\
    &\geq \frac{\widehat{P}(h)}{\widehat{P}_+(h)} - \mathcal{O}(\varepsilon(\numU, \delta_\rmU)) - \mathcal{O} \left( \varepsilon(\numP, \delta_\rmP) \right).
\end{align*}
Note that $\frac{\varepsilon(\numP, \delta_\rmP)}{\widehat{P}_+(h)} < \gamma < 1$ by the assumption on $\mathcal{H}_r$. And for the right hand side,
\begin{align*}
    \frac{\widehat{P}(h) + \varepsilon(\numU, \delta_\rmU)}{\widehat{P}_+(h) - \varepsilon(\numP, \delta_\rmP)} &= \left( \frac{\widehat{P}(h)}{\widehat{P}_+(h)} + \frac{\varepsilon(\numU, \delta_\rmU)}{\widehat{P}_+(h)} \right) \sum_{i = 0}^\infty \left( \frac{\varepsilon(\numP, \delta_\rmP)}{\widehat{P}_+(h)} \right) ^ i \\
    &= \frac{\widehat{P}(h)}{\widehat{P}_+(h)} + \frac{\varepsilon(\numU, \delta_\rmU)}{\widehat{P}_+(h)} \\
    &\ \quad + \left( \frac{\widehat{P}(h)}{\widehat{P}_+(h)} + \frac{\varepsilon(\numU, \delta_\rmU)}{\widehat{P}_+(h)} \right) \frac{\varepsilon(\numP, \delta_\rmP)}{\widehat{P}_+(h)} \sum_{i=1}^\infty \left( \frac{\varepsilon(\numP, \delta_\rmP)}{\widehat{P}_+(h)} \right)^{i-1} \\
    &\leq \frac{\widehat{P}(h)}{\widehat{P}_+(h)} + \mathcal{O}(\varepsilon(\numU, \delta_\rmU)) + \mathcal{O} \left( \varepsilon(\numP, \delta_\rmP) \right).
\end{align*}
Therefore, assign $\delta_\rmP = 1/\numP$, $\delta_\rmU = 1/\numU$ to obtain
\begin{align*}
    \sup_{h \in \mathcal{H}_r} \abs{\frac{\widehat{P}(h)}{\widehat{P}_+(h)} - \frac{P(h)}{P_+(h)}} \leq \mathcal{O} \left( \sqrt{\frac{\log\numP}{\numP}} + \sqrt{\frac{\log\numU}{\numU}} \right).
\end{align*}

\begin{figure}
    \centering
    \includegraphics[scale=0.6]{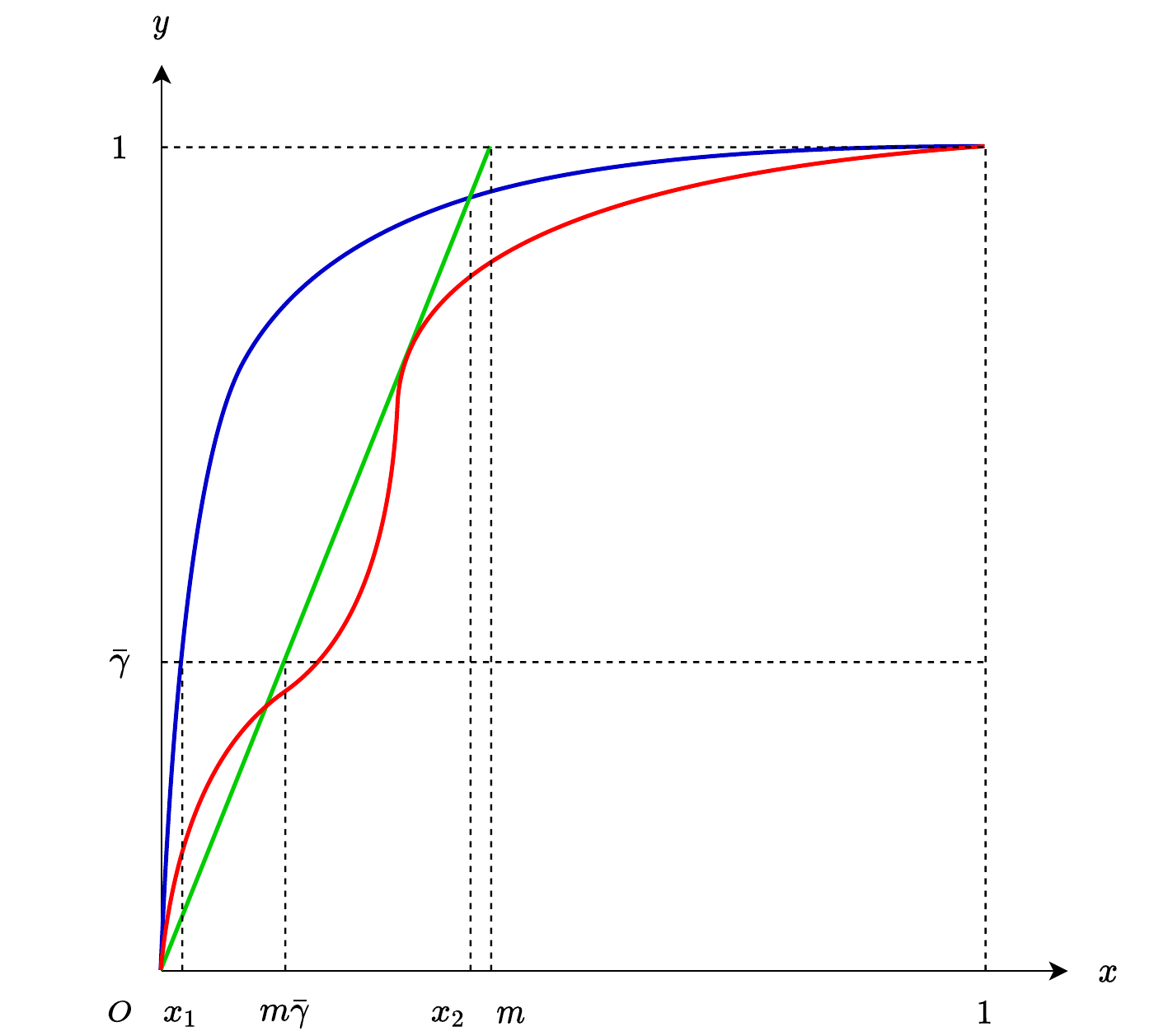}
    \caption{Green: $y = \frac{1}{m}x$ / Red: ROC curve of $r$ / Blue: ROC curve of $r^*$.}
    \label{fig: roc_curve}
\end{figure}

Next, we consider the second term of Eq. (\ref{eq: decomposition}). At first we show that the term is bounded above by $\displaystyle \frac{2(1 - \pri)}{1 - \bar{\gamma}^2} R_\AUC(r)$. Utilize $P(h) = \pri P_+(h) + (1 - \pri) P_-(h)$ to obtain
\begin{align*}
    \abs{\inf_{h \in \mathcal{H}_r} \frac{P(h)}{P_+(h)} - \pri} &= (1 - \pri) \abs{\inf_{h \in \mathcal{H}_r} \frac{P_-(h)}{P_+(h)}}.
\end{align*}
Let $\displaystyle m = \inf_{h \in \mathcal{H}_r} \frac{P_-(h)}{P_+(h)}$, and consider a ROC curve where $P_-$ (False Positive Rate) plotted as $x$-axis and $P_+$ (True Positive Rate) plotted as $y$-axis. As seen in figure \ref{fig: roc_curve}, the trapezoid surrounded by $x = 0$, $y = 1$, $y = \bar{\gamma}$, and $y = x/m$ is in the area over the ROC curve of $r$, since $\displaystyle \sup_{h \in \mathcal{H}_r} \frac{P_+(h)}{P_-(h)} = \frac{1}{m}$. Thus
\begin{align*}
    \frac{(m\bar{\gamma} + m)(1 - \bar{\gamma})}{2} &\leq 1 - \AUC(r) = R_\AUC(r) \\
    \frac{m(1 - \bar{\gamma}^2)}{2} &\leq R_\AUC(r).
\end{align*}
Then,
\begin{align*}
    \inf_{h \in \mathcal{H}_r} \frac{P_-(h)}{P_+(h)} = (1 - \pri)m \leq \frac{2(1 - \pri)}{1 - \bar{\gamma}^2} R_\AUC(r).
\end{align*}
Secondly, we prove that there exists an increasing function $\xi$ such that $m \leq \xi \left( R_\AUC(r) - R_\AUC^* \right)$. For a fixed $m$, the ROC curve of $r$ is always under $y = x/m$ in $m\bar{\gamma} \leq x \leq 1$, and the optimal ROC curve always dominates all other ROC curves \citep{Menon2016br}. Therefore, at least, the area surrounded by $y = \bar{\gamma}$, $y=x/m$, and the optimal ROC curve is assured. That is,
\begin{align*}
    R_\AUC(r) - R_\AUC^* &= \AUC(r^*) - \AUC(r) \\
    &\geq \int_{x_1}^{m\bar{\gamma}} (\rho(x) - \bar{\gamma}) \mathrm{d}x + \int_{m\bar{\gamma}}^{x_2} \left( \rho(x) - \frac{1}{m}x \right) \mathrm{d}x,
\end{align*}
where $\rho(x)$ means the optimal ROC curve and $\rho(x_1) = \bar{\gamma}$, $\rho(x_2) = x_2 / m$. Denote the right-hand side of the above inequality as $U(m)$, then
\begin{align*}
    \frac{\partial U}{\partial m}(m) &= \rho(m\bar{\gamma})\bar{\gamma} - \bar{\gamma}^2 + \rho(x_2) x_2^\prime - \rho(m\bar{\gamma})\bar{\gamma} - \frac{1}{m}x_2 x_2^\prime + \frac{x_2^2}{2m^2} + \frac{\bar{\gamma}^2}{2} \\
    &= \frac{x_2^2}{2m^2} + \frac{\bar{\gamma}^2}{2} > 0.
\end{align*}
Note that $x_1$ is independent of $m$. This shows that $U(m)$ is a strictly increasing function of $m \in [x_1 / \bar{\gamma}, 1]$, therefore, there exists an increasing function $U^{-1} : [0, 1] \to [x_1 / \bar{\gamma}, 1]$ and
\begin{align*}
    U(m) &\leq R_\AUC(r) - R_\AUC^* \\
    m &\leq U^{-1}\left( R_\AUC(r) - R_\AUC^* \right),
\end{align*}
then we can define an increasing function
\begin{align*}
    (1 - \pri)m &\leq \xi(R_\AUC(r) - R_\AUC^*) \\
    &= \min\left( \frac{2(1 - \pri)}{1 - \bar{\gamma}^2}R_\AUC(r), (1 - \pri) U^{-1}\left( R_\AUC(r) - R_\AUC^* \right) \right).
\end{align*}
Thirdly, we check that $\xi(0) \to 0$ when $\bar{\gamma} \to 0$. Since the optimal ROC curve is concave \citep{Menon2016br}, we have $\displaystyle U(x_1 / \bar{\gamma}) = 0$. And from the irreducibility assumption, we have $\displaystyle \frac{x_1}{\bar{\gamma}} \to 0$ when $\bar{\gamma} \to 0$. Therefore, $\displaystyle U^{-1}(0) = \frac{x_1}{\bar{\gamma}} \to 0$.
and we concludes the proof. \hfill $\qedsymbol$


\subsection{Proof of Proposition \ref{thm: generator_of_BR}}
From \citet{Reid2009}, we have the integral representation of the Bregman divergence:
\begin{align*}
    \BR{f}{r^*}{r} = \expectm{X}{\int_{r(X)}^{r^*(X)} (r^*(X) - t) \ddi{f}(t) \mathrm{d}t}.
\end{align*}
Then,
\begin{align*}
    \BR{f}{r^*}{r} &= \expectm{X}{\int_r^{r^*} (r^* - t) \ddi{f}(t) \mathrm{d}t} \\
    &\geq \expectm{X}{\int_r^{r^*} (r^* - t) \mu \mathrm{d}t} \\
    &= \BR{f_\mathrm{S}}{r^*}{r}.
\end{align*}
We used $\ddi{f}(t) \geq \inf_{t \in [0, \infty)} \ddi{f}(t) = \mu$ at the second line and $\ddi{f_\mathrm{S}}(t) = \mu$ at the third line. \hfill $\qedsymbol$


\subsection{Proof of Proposition \ref{thm: squared_loss}}
As mentioned in the Proof of Proposition \ref{thm: generator_of_BR}, we have the integral representation of the Bregman divergence. Then,
\begin{align*}
    \BR{f_\mathrm{S}}{r^*}{r} &= \expectm{X}{\int_{r(X)}^{r^*(X)} (r^*(X) - t) \ddi{f_\mathrm{S}}(t) \mathrm{d}t} \\
    &= \mu \expectm{X}{\int_{r^*}^r (t - r^*) \mathrm{d}t} \\
    &= \mu \expectm{X}{\int_{r^*}^{\min(r, 1 / \pri)} (t - r^*) \mathrm{d}t + \indi{\pri r > 1} \int_{1/\pri}^r (t - r^*) \mathrm{d}t}.
\end{align*}
The first term in the expectation can be written as:
\begin{align*}
    \int_{r^*}^{\min(r, 1 / \pri)} (t - r^*) \mathrm{d}t &= \frac{1}{\pri^2} \int_{\eta}^{\min(\pri r, 1)} (t - \eta) \mathrm{d}t,
\end{align*}
where $\eta(x) = \pri r^*(x) = P(Y = +1 \mid X = x)$. 
On the other hand, the classification risk~w.r.t the squared loss can be written as
\begin{align*}
    R_\mathrm{sq}(g) &= \expectm{X, Y}{\frac{1}{4}(Yg(X) - 1)^2} \\
    &= \expectm{X}{C_\eta(g)}
\end{align*}
where $g \in [-1, 1]$ and $C_\eta(g)$ is the conditional risk
\begin{align*}
    C_\eta(g) = \eta \frac{(g - 1)^2}{4} + (1 - \eta) \frac{(g + 1)^2}{4},
\end{align*}
and the optimal classification risk~w.r.t the squared loss as
\begin{align*}
    C_\eta^* = \inf_g C_\eta(g) = C_\eta(2\eta - 1).
\end{align*}
Then, we have
\begin{align*}
    C_\eta(g) - C_\eta^* &= \eta (\min(\pri r, 1) - 1)^2 + (1 - \eta)(\min(\pri r, 1))^2 - \eta(1 - \eta) \\
    &= (\min(\pri r, 1) - \eta)^2 \\
    &= 2 \int_\eta^{\min(\pri r, 1)} (t - \eta) \mathrm{d}t.
\end{align*}
Thus,
\begin{align*}
    R_\mathrm{sq}(g) - R_\mathrm{sq}^* &= \expectm{X}{C_\eta(g) - C_\eta^*} \\
    &= 2 \expectm{X}{\int_\eta^{\min(\pri r, 1)} (t - \eta) \mathrm{d}t.}
\end{align*}
Next, the second term in the expectation is 
\begin{align*}
    \indi{\pri r > 1} \int_{1/\pri}^r (t - r^*) \mathrm{d}t = \indi{\pri r > 1} \cdot \frac{1}{2}(\pri r - 1)(\pri r - 2\eta + 2),
\end{align*}
then,
\begin{align*}
    &\ \expectm{X}{\indi{\pri r > 1} \int_{1/\pri}^r (t - r^*) \mathrm{d}t} \\ 
    =&\ \frac{1}{2} \expectm{X \mid \pri r(X) > 1}{(\pri r - 1)(\pri r - 2\eta + 2)} P(\pri r(X) > 1).
\end{align*}
Finally, we have
\begin{align*}
    \frac{2 \pri^2}{\mu} \BR{f_\mathrm{S}}{r^*}{r} = R_\mathrm{sq}(g) - R_\mathrm{sq}^* + \chi_r \mathfrak{D}_r.
\end{align*}
\hfill $\qedsymbol$


\subsection{Proof of Theorem \ref{thm: excess_auc}}
From the result of \citet{Clemencon2006}, we have
\begin{align*}
    &\ R_\AUC(s) - R_\AUC^* \\
    =&\ \frac{1}{2\pri(1 - \pri)}\expectm{X, X^\prime}{\abs{\eta(X) - \eta(X^\prime)} \indi{(s(X) - s(X^\prime)(\eta(X) - \eta(X^\prime)) \leq 0}},
\end{align*}
where $\eta(x) = p(Y = +1 \mid X = x)$. Then for any $r$,
\begin{align*}
    &\ R_\AUC(r) - R_\AUC^* \\
    =&\ \frac{1}{2\pri(1 - \pri)} \expectm{X, X^\prime}{\abs{\eta(X) - \eta(X^\prime)} \indi{(r(X) - r(X^\prime))(\eta(X) - \eta(X^\prime)) \leq 0}} \\
    =&\ \frac{1}{2\pri(1 - \pri)} \expectm{X, X^\prime}{\abs{\eta(X) - \eta(X^\prime)} \indi{(r(X) - r(X^\prime))(\eta(X) - \eta(X^\prime)) \leq 0}} \\
    =&\ \frac{1}{2(1 - \pri)} \mathbb{E}_{X, X^\prime} \Big[ \abs{r^*(X) - r^*(X^\prime)} \Big( \indi{r(X) < r(X^\prime)} \indi{r^*(X) \geq r^*(X^\prime)} \\
    &\ \qquad \qquad \qquad \quad + \indi{r(X) \geq r(X^\prime)} \indi{r^*(X) < r^*(X^\prime)} \Big) \Big] \\
    =&\ \frac{1}{2(1 - \pri)} \mathbb{E}_{X, X^\prime} \Big[ (r^*(X) - r^*(X^\prime))\indi{r(X) < r(X^\prime)} \\
    &\ \qquad \qquad \qquad \quad + (r^*(X^\prime) - r^*(X))\indi{r(X) \geq r(X^\prime)} \Big] \\
    \leq&\ \frac{1}{2(1 - \pri)} \mathbb{E}_{X, X^\prime} \big[ \left(r^*(X) - r(X) + r(X^\prime) - r^*(X^\prime) \right) \indi{r(X) < r(X^\prime)} \\
    &\ \qquad \qquad \qquad \quad + \left( r^*(X^\prime) - r(X^\prime) + r(X) - r^*(X) \right) \indi{r(X) \geq r(X^\prime)} \big] \\
    \leq&\ \frac{1}{2(1 - \pri)} \expectm{X, X^\prime}{\abs{r^*(X) - r(X)} + \abs{r^*(X^\prime) - r(X^\prime)}} \\
    =&\ \frac{1}{1 - \pri} \expectm{X}{\abs{r^*(X) - r(X)}} \\
    \leq&\ \frac{1}{1 - \pri} \sqrt{\frac{2}{\mu} \BR{f}{r^*}{r}}.
\end{align*}
Used the same technique as the proof of Theorem \ref{thm: classification_ratio} for the last inequality, and this concludes the proof.
\hfill $\qedsymbol$


\subsection{Proof of Theorem \ref{thm: classification_threshold}}
We utilize the following equality (see the proof of Theorem \ref{thm: classification_shift}.)
\begin{align*}
    R_{\pris, \costs}(h_{\hat{\theta}}) - R_{\pris, \costs}^* &= \frac{\costs + \pris - 2\costs \pris}{\cost_0 + \pri - 2\cost_0 \pri} (R_{\pri, \cost_0}(h_{\hat{\theta}}) - R_{\pri, \cost_0}^*) \\
    &= C \left( \left( R_{\pri, \cost_0}(h_{\hat{\theta}}) - R_{\pri, \cost_0}(h_{\cost_0}) \right) + \left( R_{\pri, \cost_0}(h_{\cost_0}) - R_{\pri, \cost_0}^* \right) \right),
\end{align*}
where $C = \frac{\costs + \pris - 2\costs \pris}{\cost_0 + \pri - 2\cost_0 \pri}$ and $h_{\cost_0} = \sign(\pri r - \cost_0) = \sign(r - \theta)$. The second term $R_{\pri, \cost_0}(h_{\cost_0}) - R_{\pri, \cost_0}^*$ is processed by Theorem \ref{thm: classification_ratio}, so we consider the first term. By the definition of cost-sensitive classification risk,
\begin{align*}
    &\ R_{\pri, \cost_0}(h_{\hat{\theta}}) - R_{\pri, \cost_0}(h_{\cost_0}) \\ 
    =&\ \mathbb{E}_X \Big[ (1 - \cost_0)\indi{Y=+1} \left( \indi{\sign(r(X)-\hat{\theta})=-1} - \indi{\sign(r(X)-\theta)=-1} \right) \\
    & \quad \quad + \cost_0 \indi{Y=-1} \left( \indi{\sign(r(X)-\hat{\theta})=+1} - \indi{\sign(r(X)-\theta)=+1} \right) \Big] \\
    =&\ \mathbb{E}_X \Big[ (1 - \cost_0) \eta(X) \left( \indi{\theta < r(X) < \hat{\theta}} - \indi{\hat{\theta} < r(X) < \theta} \right) \\
    &\quad \quad + \cost_0 (1 - \eta(X)) \left( \indi{\hat{\theta} < r(X) < \theta} - \indi{\theta < r(X) < \hat{\theta}} \right) \Big] \\
    =&\ \expectm{X}{(\eta(X) - \cost_0) \left( \indi{\theta < r(X) < \hat{\theta}} - \indi{\hat{\theta} < r(X) < \theta} \right)} \\
    =&\ \pri \expectm{X}{(r^*(X) - \theta) \left( \indi{\theta < r(X) < \hat{\theta}} - \indi{\hat{\theta} < r(X) < \theta} \right)} \\
    =&\ \pri \expectm{X}{(r^*(X) - r(X) + r(X) - \theta) \left( \indi{\theta < r(X) < \hat{\theta}} - \indi{\hat{\theta} < r(X) < \theta} \right)} \\
    \leq&\ \pri \omega_{\hat{\theta}} \expectm{X}{\abs{r^*(X) - r(X)}} + \pri \abs{\hat{\theta} - \theta} \\
    \leq&\ \pri \omega_{\hat{\theta}} \sqrt{\frac{2}{\mu} \BR{f}{r^*}{r}} + \pri \abs{\hat{\theta} - \theta}.
\end{align*}
The first inequality holds because $\expect{\abs{\indi{\theta < r(X) < \hat{\theta}} - \indi{\hat{\theta} < r(X) < \theta}}} \leq 1$ and $\expect{\indi{\theta < r(X) < \theta}} = 0$. We used the same technique as the proof of Theorem \ref{thm: classification_ratio} for the second inequality, and this concludes the proof.
\hfill $\qedsymbol$


\subsection{Proof of Proposition \ref{thm: threshold_error}}
We first consider the bound for $\abs{\hat{\theta} - \theta}$. Let $\hat{\pri} = \pri + d$, $\hat{\pri}^\prime = \pris + d^\prime$ where $-\pri < d < 1 - \pri$ and $-\pris < d^\prime < 1 - \pris$. Then, 
\begin{align*}
    \hat{\theta} &= \frac{\costs (1 - \hat{\pri}^\prime)}{(1 - \costs)(1 - \hat{\pri})\hat{\pri}^\prime + \costs \hat{\pri}(1 - \hat{\pri}^\prime)} \\
    &= \frac{\costs(1 - \pris) - \costs d^\prime}{(1 - \costs)(1 - \pri)\pris + \costs \pri (1 - \pris) + (\costs - \pris)d + (1 - \costs - \pri) d^\prime - dd^\prime}.
\end{align*}
We denote $A = \costs(1 - \pris)$, $B(d^\prime) = -\costs d^\prime$, $C = (1 - \costs)(1 - \pri)\pris + \costs \pri (1 - \pris)$, $D(d, d^\prime) = (\costs - \pris)d + (1 - \costs - \pri)d^\prime - dd^\prime$. Using these notations, we have
\begin{align*}
    \abs{\hat{\theta} - \theta} &= \abs{\frac{A + B}{C + D} - \frac{A}{C}}.
\end{align*}
Thus, if $D \geq 0$, 
\begin{align*}
    \abs{\hat{\theta} - \theta} &= \abs{\frac{A + B}{C + D} - \frac{A}{C}} \\
    &= \abs{\frac{A + B}{C + D} - \frac{A}{C + D} \frac{C + D}{C}} \\
    &= \abs{\frac{A}{C + D} + \frac{B}{C + D} - \frac{A}{C + D} \left( 1 + \frac{D}{C} \right)} \\
    &= \abs{\frac{1}{C + D} \left( B - \frac{AD}{C} \right)} \\
    &\leq \frac{1}{C} \abs{B - \frac{AD}{C}} \\
    &\leq \mathcal{O}\left( \abs{d} + \abs{d^\prime} \right) \quad (\mathrm{as} \ d, d^\prime \to 0),
\end{align*}
where the first inequality holds because $C > 0$ and $D > 0$. And if $D < 0$, we have $C > D$ from $\hat{\theta} \geq 0$. Then,
\begin{align*}
    \abs{\hat{\theta} - \theta} &= \abs{\frac{A + B}{C + D} - \frac{A}{C}} \\
    &= \abs{\frac{A + B}{C} \sum_{i=0}^\infty \left( -\frac{D}{C} \right)^i - \frac{A}{C}} \\
    &= \abs{\frac{A + B}{C} + \frac{A + B}{C} \sum_{i=1}^\infty \left( -\frac{D}{C} \right)^i - \frac{A}{C} } \\
    &= \frac{1}{C} \abs{B + (A + B) \sum_{i=1}^\infty \left( -\frac{D}{C} \right)^i} \\
    &\leq \mathcal{O}\left( \abs{d} + \abs{d^\prime} \right) \quad (\mathrm{as} \ d, d^\prime \to 0).
\end{align*}
and we complete the proof.


\subsection{Proof of Corollary \ref{thm: convergence}}
It immediately holds from Theorem 1 of \citet{Kato2020} and Theorem \ref{thm: classification_threshold}.
\hfill $\qedsymbol$


%% file: declarations.tex

\subsection*{Funding}
Masashi Sugiyama was supported by KAKENHI 20H04206.

\subsection*{Conflicts of interest/Competing interests}
Not Applicable.

\subsection*{Ethics approval}
Not Applicable.

\subsection*{Consent to participate}
Not Applicable.

\subsection*{Consent for publication}
Not Applicable.

\subsection*{Availability of data and material}
Only public datasets and frameworks were used.

\subsection*{Code availability}
Our code is available at \url{https://github.com/csnakajima/pu-learning}.

\subsection*{Authors' contributions}
All authors contributed to the study conception and design. Theoretical analysis and experimental setup were performed by Shota Nakajima. The first draft of the manuscript was written by Shota Nakajima, and all authors commented on previous versions of the manuscript. All authors read and approved the final manuscript.